%% file: paper.tex
\documentclass[]{jingdong}
\usepackage[toc,page,header]{appendix}
\input{common}

\usepackage{cleveref}
\theoremstyle{plain}

\theoremstyle{definition}

\theoremstyle{remark}
\usepackage{subcaption}

\usepackage[textsize=tiny]{todonotes}
\usepackage{fontawesome}

\usepackage{xspace}
\usepackage{array}
\usepackage[most]{tcolorbox}
\usepackage{wrapfig}
\usepackage{threeparttable}
\usepackage{algorithm}
\usepackage[ruled,vlined,noend,algo2e]{algorithm2e}
\usepackage{multirow}
\usepackage{tabularx}
\usepackage{makecell}
\usepackage{bbm}
\usepackage{bbding}
\usepackage{siunitx}
\usepackage{ulem}
\usepackage{arydshln}

\newcommand{\ourmethod}{{\fontfamily{lmtt}\selectfont \textbf{TodoEvolve}}\xspace}
\newcommand{\ourmodel}{{\fontfamily{lmtt}\selectfont \textbf{Todo-14B}}\xspace}
\newcommand{\ourframework}{{\fontfamily{lmtt}\selectfont \textbf{PlanFactory}}\xspace}

\newcommand{\hlblock}[1]{%
  \setlength\fboxsep{1pt}%
  \colorbox[HTML]{E39963}{\textcolor{white}{#1}}%
}

\usepackage{libertine}

\definecolor{tblheader}{gray}{0.95}
\definecolor{symblue}{RGB}{25, 100, 180}
\definecolor{symred}{RGB}{180, 50, 50}
\definecolor{peach}{RGB}{255,218,185}

\newcolumntype{L}[1]{>{\raggedright\arraybackslash}p{#1}}
\newcolumntype{C}[1]{>{\centering\arraybackslash}p{#1}}

\title{TodoEvolve: Learning to Architect Agent Planning Systems}

\affiliation{TodoRL Team}

\abstract{

Planning has become a central capability for contemporary agent systems in navigating complex, long-horizon tasks, yet existing approaches predominantly rely on fixed, hand-crafted planning structures that lack the flexibility to adapt to the structural diversity of open-ended problems. To address this limitation, we introduce \ourmethod, a meta-planning paradigm that autonomously synthesizes and dynamically revises task-specific planning architectures.
Specifically, we first construct \ourframework, a modular design space that standardizes diverse planning paradigms within a unified codebase encompassing topology, initialization, adaptation, and navigation, thereby providing a common interface for heterogeneous planning patterns. Leveraging \ourframework, we collect high-quality planning trajectories and train \ourmodel via \textit{Impedance-Guided Preference Optimization} (IGPO), a multi-objective reinforcement learning objective that encourages the generation of planning systems that are performant, stable, and token-efficient across arbitrary tasks and agent backbones.
Empirical evaluations on five agentic benchmarks demonstrate that \ourmethod consistently surpasses carefully engineered planning modules while maintaining economical API costs and runtime overhead.
}

\date{\today}
\checkdata[Code]{\url{https://github.com/EcthelionLiu/TodoEvolve}}

\begin{document}
\maketitle

\section{Introduction}
With the rapid advancement of foundation models~\citep{kimiteam2025kimik2openagentic,5team2025glm45agenticreasoningcoding,tongyideepresearchteam2025tongyideepresearchtechnicalreport}, large language model (LLM)-powered agents have begun to demonstrate strong capabilities across domains such as deep research~\citep{hu2025stepdeepresearchtechnicalreport,shi2025deepresearchsystematicsurvey}, complex software engineering~\citep{iquestlabIQuestCoder,yang2024sweagentagentcomputerinterfacesenable}, and real-world transactions~\cite{andonlabsVendingBenchAndon,backlund2025vendingbenchbenchmarklongtermcoherence}. Beyond improvements in base model capacity, increasingly sophisticated agent scaffolds are equally critical~\citep{wang2025openhandsopenplatformai}, equipping LLMs with essential agentic support including planning~\citep{parmar2025plangenmultiagentframeworkgenerating,wu2025gapgraphbasedagentplanning,erdogan2025planandactimprovingplanningagents}, memory~\citep{hu2026memoryageaiagents}, reflection, \textit{etc}. Among these, planning stands out as a central capability, enabling agents to navigate complex environments by maintaining a coherent global state, preserving behavioral consistency, and coordinating actions across tasks~\citep{cao2025largelanguagemodelsplanning}.

Existing planning systems developed for LLM-based agents exhibit substantial diversity. From the perspective of \textbf{planning target}, some are designed to support \textit{single agent}, primarily addressing long-horizon execution and mitigating the risk of ``\textit{lost in the middle}''~\citep{erdogan2025plan-and-act}, while others are tailored for \textit{multi-agent systems}, focusing on subtask allocation and contextual coordination across agents with distinct roles~\citep{parmar2025plangenmultiagentframeworkgenerating,hu2025owloptimizedworkforcelearning}. In terms of \textbf{representational form}, plans have been instantiated using a wide range of structures, including linear to-do lists~\citep{githubGitHubLangchainaideepagents}, directed acyclic graphs (DAG)~\citep{qin2025flashsearcherfasteffectiveweb}, tree-structured plans~\citep{hu2026flowsearchadvancingdeepresearch}, and hierarchical notes. Moreover, planning systems differ markedly across \textbf{task domains}, with domain-specific designs emerging for embodied action~\citep{wang2024describeexplainplanselect}, web search~\citep{kim-etal-2024-rada}, and programming. Faced with this diversity, practitioners may naturally ask: \textit{is there a single planning structure that can serve as a \emph{one-size-fits-all} solution that generalizes well across settings?}

We posit that such an oracle planning system does not exist. Beyond distinct task domains require different planning priors (for instance, MCTS-based planning may be effective for mathematical reasoning yet is rarely adopted for autonomous driving agents due to the vastness of its action space~\citep{wang2024qimprovingmultistepreasoning}), even within a single task class, alternative planning priors exhibit performance disparities. For example, in web search, AOP~\citep{li2025agentorientedplanningmultiagentsystems} employs a simple linear to-do list coupled with a reward model to solve document QA in a token-efficient manner, but it is substantially outperformed in more complex multimodal settings by DAG-based planning structures~\citep{qin2025flashsearcherfasteffectiveweb}. Similarly, while linear tasks require minimal revision~\citep{hu2025owloptimizedworkforcelearning}, high-conflict environments demand continuous topological restructuring~\citep{zhang2025cosightenhancingllmbasedagents}, rendering a single, universal planning system unrealistic.

Accordingly, we contend that the central challenge is not to design a \textit{one-size-fits-all} planner, but to \textit{customize} planning systems to the structural characteristics of each task. To this end, we propose \ourmethod, a meta-planning paradigm that synthesizes task-adaptive agentic planners and dynamically updates their planning states as execution unfolds. Concretely, we train \ourmodel using \textit{Impedance-Guided Preference Optimization} (IGPO), a multi-objective preference learning objective that jointly promotes high performance, stability, and token efficiency in the generated planning systems. The resulting meta-planner \ourmodel takes a task instance as input and instantiates a tailored planning topology, revision cadence, and navigation strategy, operationalized as a task-specific \textit{to-do} structure. \ourmodel integrates seamlessly with single/multi-agent execution frameworks, remains compatible with diverse LLM backbones, and generalizes across heterogeneous task domains.

To ground \ourmethod within the diverse landscape of existing planning systems, we introduce a modular \emph{planning design space} comprising four dimensions: \ding{168} \textbf{Topology} (the structural organization of task decomposition), \ding{169} \textbf{Initialization} (how the task topology is instantiated), \ding{170} \textbf{Adaptation} (when and how the topology is revised), and \ding{171} \textbf{Navigation} (the mechanism that issues executable directives to the acting agent). This design space provides a unified abstraction capable of accommodating and localizing a wide spectrum of existing planning paradigms.  
Building on this formulation, we decompose and re-implement ten representative planning architectures, including Plan-and-Act~\citep{erdogan2025plan-and-act}, linear planning~\citep{hu2025owloptimizedworkforcelearning}, DAG-based planning~\citep{qin2025flashsearcherfasteffectiveweb}, and parallel and dynamic planning~\citep{zhu2025oagentsempiricalstudybuilding}. The resulting framework, denoted as \ourframework, serves both as \textbf{(i) a data synthesis engine} for generating high-quality planning trajectories to train \ourmethod and \textbf{(ii) a standardized codebase} to facilitate future research on agentic planning capabilities. Our contributions are as follows:

\vspace{-1pt}
\begin{itemize}[leftmargin=1.2em,itemsep=-0.2em]
\item[\ding{182}] \textbf{Unified Codebase:} We introduce \ourframework, a modular design space for agentic planning systems encompassing four key components (\textit{topology}, \textit{initialization}, \textit{adaptation}, and \textit{navigation}), providing unified implementations and benchmark support for a wide range of prevailing planning structues.

\item[\ding{183}] \textbf{Meta Planners:} We introduce \ourmethod, a meta-planning paradigm that synthesizes task-adaptive planning systems and dynamically revises planning states. Through impedance-guided preference optimization (IGPO), we train \ourmodel, a meta-planner capable of instantiating and controlling planning structures across diverse scenarios and agent backbones.

\item[\ding{184}] \textbf{Experimental Evaluation:} Extensive experiments on four challenging agentic benchmarks demonstrate that \ourmethod delivers (I) \textbf{substantial performance gains}, improving frameworks such as Smolagents by up to {$16.37\%$} on GAIA; and (II) \textbf{robust generalization}, generalizing across diverse LLM backbones, for example boosting GPT-5-Mini to $75\%$ on xBench-DS.
\end{itemize}

\vspace{-1pt}
\section{Related Works}

\definecolor{tblheader}{gray}{0.95} 
\definecolor{symblue}{RGB}{25, 100, 180} 
\definecolor{symred}{RGB}{180, 50, 50}   
\definecolor{sympurp}{RGB}{130, 50, 130} 
\begin{table*}[t]
\centering
\small
\caption{An overview of agentic planning paradigms decomposed in \ourframework. 
The ``Mul'' column distinguishes between single-agent ($\mathcal{S}$) and multi-agent ($\mathcal{M}$) compatibility. ``Scope'' specifies the granularity at which planning is performed (\textcolor{symred}{$\alpha$} for step-wise vs. \textcolor{symblue}{$\Omega$} for task-wise), and ``Struct'' indicates whether the execution flow is linear ($\ell$) or organized as a complex graph structure ($\mathcal{G}$).}
\label{tab:plan_system_refined}

\renewcommand{\arraystretch}{1.5} 
\setlength{\tabcolsep}{4pt}       

\resizebox{\textwidth}{!}{%
\begin{tabular}{l l c c c L{3.0cm} L{3.0cm} L{3.0cm} L{3.0cm}}
\toprule
\rowcolor{tblheader} 
 & & \textbf{Mul.} & \textbf{Scope} & \textbf{Struct.} & 
\textcolor{symblue}{\ding{168}} \textbf{Topology} & 
\textcolor{symblue}{\ding{169}} \textbf{Initialization} & 
\textcolor{symblue}{\ding{170}} \textbf{Adaptation} & 
\textcolor{symblue}{\ding{171}} \textbf{Navigation} \\

\rowcolor{tblheader}
\multirow{-2}{*}{\textbf{Method}} & \multirow{-2}{*}{\textbf{Date}} & 
\scriptsize ($\mathcal{M}/\mathcal{S}$) & \scriptsize ($\Omega/\alpha$) & \scriptsize ($\mathcal{G}/\ell$) & 
\scriptsize \textit{Structural Organization} & 
\scriptsize \textit{Instantiation Mechanism} & 
\scriptsize \textit{Revision Logic} & 
\scriptsize \textit{Execution Directives} \\
\midrule

\textbf{OWL} & 2025.6 & $\mathcal{M}$ & \textcolor{symblue}{$\Omega$} & $\mathcal{G}$ & 
Dual Hierarchy & 
Planner Decompose & 
Manager Intervention & 
Dynamic Dispatch \\

\textbf{OAgents} & 2025.6 & $\mathcal{M}$ & \textcolor{symred}{$\alpha$} & $\ell$ & 
Modular Graph & 
SOP Configuration & 
Critic-Loop Feedback & 
Loop Execution \\

\textbf{AgentOrchestra} & 2025.9 & $\mathcal{M}$ & \textcolor{symblue}{$\Omega$} & $\mathcal{G}$ & 
Orch. Hierarchy & 
Role Definition & 
Env Feedback & 
Centralized Routing \\

\textbf{Flash-Searcher} & 2025.9 & $\mathcal{S}$ & \textcolor{symblue}{$\Omega$} & $\mathcal{G}$ & 
Parallel DAG & 
Dependency Parsing & 
Workflow Pruning & 
Concurrent Paths \\

\textbf{JoyAgent} & 2025.10 & $\mathcal{M}$ & \textcolor{symblue}{$\Omega$} & $\mathcal{G}$ & 
Collective Hierarchy & 
Hybrid Planning & 
Consensus Voting & 
Joint Deliberation \\

\textbf{FlowSearch} & 2025.10 & $\mathcal{M}$ & \textcolor{symblue}{$\Omega$} & $\mathcal{G}$ & 
Thought Graph & 
Flow Construction & 
Dynamic Expansion & 
Graph Traversal \\

\textbf{Co-Sight} & 2025.10 & $\mathcal{M}$ & \textcolor{symred}{$\alpha$} & $\ell$ & 
Cross-Check Net & 
Inconsistency Trigger & 
Meta-Verification & 
Conflict Resolution \\

\bottomrule
\end{tabular}%
}
\vspace{-0.4em}
\end{table*}

\paragraph{Agent Planning Systems.}
Agentic planning has evolved from static prompting to structured reasoning. Foundational works like CoT~\citep{cot}, ToT~\citep{tot}, and GoT~\citep{got} enabled cognitive decomposition, while ReAct~\citep{yao2023react} and Reflexion~\citep{reflexion} introduced execution loops with self-correction. However, these approaches typically rely on rigid, predetermined topologies, limiting adaptability in open-ended environments where optimal structures vary dynamically.
Recent frameworks address this by embedding domain priors: Flash-Searcher~\citep{qin2025flashsearcherfasteffectiveweb} and OAgents~\citep{zhu2025oagentsempiricalstudybuilding} leverage DAG-based parallelism; OWL~\citep{hu2025owloptimizedworkforcelearning} and AgentOrchestra~\citep{li2025agentorientedplanningmultiagentsystems} utilize hierarchical coordination; and systems like FlowSearch~\citep{hu2026flowsearchadvancingdeepresearch}, JoyAgent~\citep{han2025joyagents}, and Co-Sight~\citep{zhang2025cosightenhancingllmbasedagents} optimize workflows via structured verification. Crucially, these systems remain bound by \textit{pre-designed} architectures. This necessitates a meta-planning approach capable of autonomously synthesizing and \emph{customizing} planning structures tailored to each task's unique complexity.

\vspace{-10pt}
\paragraph{RL for Agent Planning.}
Training paradigms have shifted from preference alignment~\citep{rafailov2023dpo,schulman2017ppo} toward reinforcement learning with verifiable rewards (RLVR)~\citep{guo2025deepseek-r1}, optimizing against objective ground truths fosters emergent self-verification.
Recent works apply this to diverse dimensions: Search-R1~\citep{jin2025search} and LATS~\citep{zhou2023language} optimize search trajectories; RAGEN~\citep{wang2025ragen} targets multi-turn interactions; and ToRL~\citep{li2025torl} refines tool-use strategies. More related works include \citep{li2025encouraginggoodprocessesneed,xi2025agentgymrltrainingllmagents,feng2024agilenovelreinforcementlearning,paglieri2025learning}. However, a critical limitation persists: these approaches primarily optimize the agent's action policy or tool selection within fixed topological loops. In contrast, our work leverages verifiable trajectories to train a meta-planner, moving beyond policy optimization to autonomously synthesize the underlying planning structure itself.

\vspace{-1pt}
\section{PlanFactory: Unified Planning Codebase}

\subsection{Preliminary}

We adopt a bi-level agentic inference abstraction where the Agent System executes environment interactions, while the Planning System governs high-level control logic.

\vspace{-10pt}
\paragraph{Agent Systems.}
We formalize the execution substrate as a tuple $\mathcal{M} = \langle I, \mathcal{S}, \mathcal{A}, \Psi, \Omega \rangle$, comprising an agent roster $I$, a global state space $\mathcal{S}$, and a joint action space $\mathcal{A} = \bigcup_{i \in I} \mathcal{A}_i$. The state dynamics follow $\Psi(s_{t+1} \mid s_t, a_t, \mu(t))$, where $\mu(t) \in I$ identifies the active agent at time $t$. To support action generation, a context mechanism $\Omega$ aggregates the execution history $H_t$, such that $a_t = \pi_{\mu(t)}(s_t, H_t, Q \mid \Omega)$. Finally, the resulting trajectory $\tau$ is evaluated by a reward $R(\tau)$, positioning $\mathcal{M}$ as a flexible execution engine orchestrated by higher-level logic.

\vspace{-10pt}
\paragraph{Planning Systems.}
The Planning System imposes structural logic on execution. We formalize it as a configuration $\mathcal{P}$ comprising four key functional modules:
\begin{equation}
    \mathcal{P} = \langle \mathcal{G}, \mathcal{I}_{init}, \mathcal{F}_{adapt}, \mathcal{N}_{nav} \rangle
\end{equation}
defining the mechanisms respectively. As shown in Table~\ref{tab:plan_system_refined}, existing paradigms represent static instances of $\mathcal{P}$, augmenting the policy as $a_t = \pi(\cdot \mid \mathcal{P})$. Crucially, current systems rely on manual engineering to fix $\mathcal{P}$, limiting adaptability. This motivates our meta-level framework, which automatically synthesizes an optimal $\mathcal{P}^*$ tailored to each task.

\vspace{-1pt}
\begin{figure*}[t]
    \centering
    \includegraphics[width=0.95\textwidth]{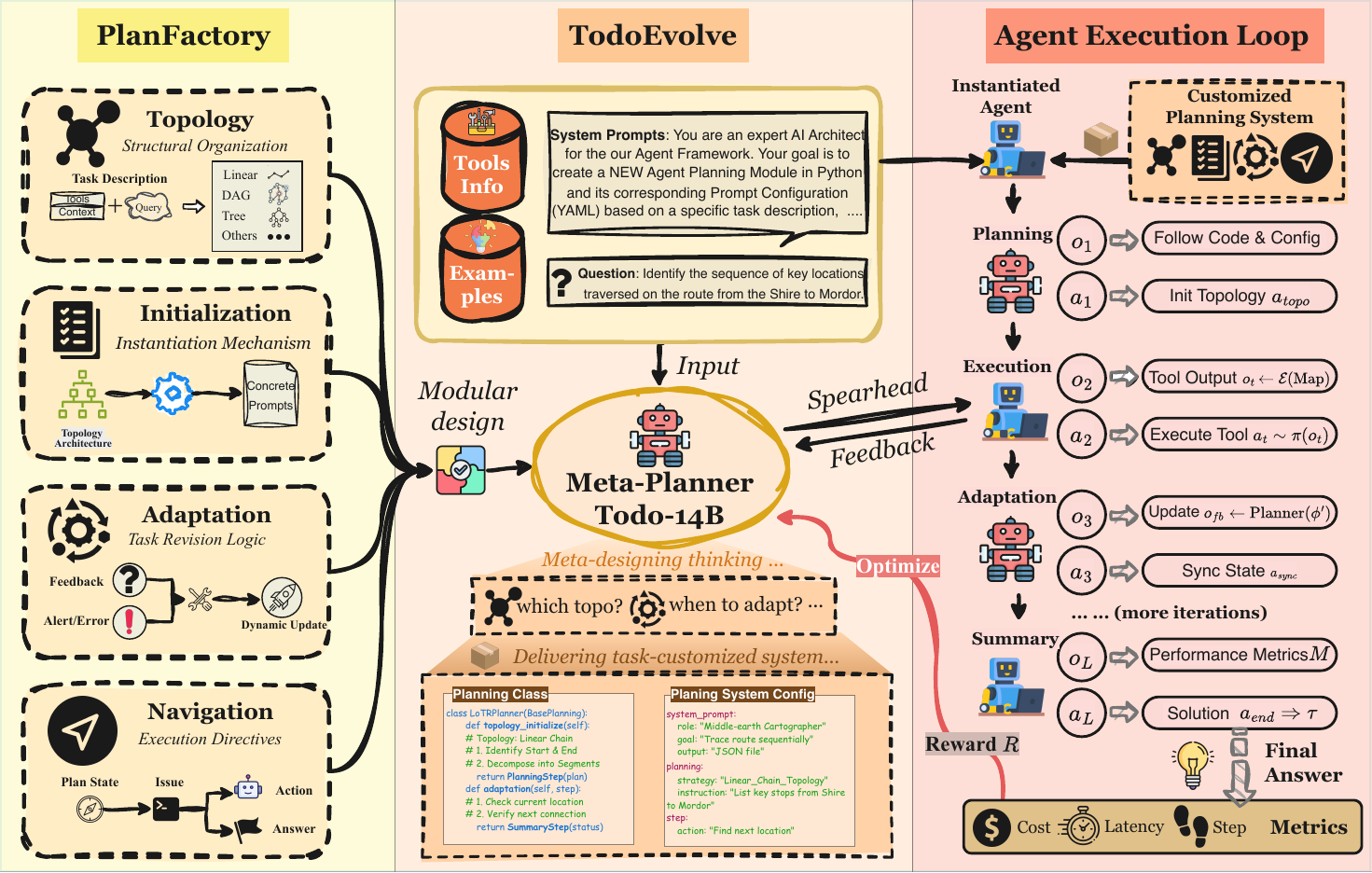}
    \caption{The overall inference workflow of \ourmethod first constructs a customized planning system along four dimensions—topology, initialization, adaptation, and navigation, and then deploys it in real time to orchestrate agent execution.}
    \label{fig:framework}
    \vspace{-0.1em}
\end{figure*}

\subsection{PlanFactory Codebase}
We present \ourframework, a modular toolkit designed to decouple high-level planning logic from low-level execution, facilitating the systematic study of agentic architectures.

\vspace{-10pt}
\paragraph{Implementation.}
The core of PlanFactory is a standardized lifecycle interface. All planning paradigms (Table~\ref{tab:plan_system_refined}) inherit from the \texttt{BasePlanning} abstract class, which encapsulates the four essential components: \ding{168}~\textbf{Topology}, \ding{169}~\textbf{Initialization}, \ding{170}~\textbf{Adaptation}, and \ding{171}~\textbf{Navigation}. For more details, please refer to Appendix~\ref{app:factory}. . This polymorphism allows heterogeneous strategies to be swapped seamlessly within a shared runtime. Crucially, this design supports highly parallelized inference, enabling users to benchmark disparate configurations concurrently on a unified backend without refactoring the agent loop.

\vspace{-10pt}
\paragraph{Evaluation.}
PlanFactory provides a comprehensive evaluation suite tailored for dynamic information-seeking tasks. To ensure reliable assessment in open domains, we employ an LLM-as-a-Judge mechanism. This automates trajectory analysis, rigorously quantifying both task success rates and the logical coherence of the generated plans.

\section{TodoEvolve: Training Meta-Planners}

Current agentic systems predominantly rely on static protocols, which inherently lack the flexibility to address the diverse  distribution of real-world queries. To break the shackles of manual engineering, we propose a Generative Planning Paradigm. The core of this paradigm is \textbf{Impedance-Guided Preference Optimization (IGPO)}, a novel training strategy designed to endue \ourmodel with the ability to dynamically synthesize bespoke planning systems $\mathcal{P}_{custom}$ tailored to unique structural requirements. Unlike standard alignment which focuses on stylistic imitation, IGPO explicitly optimizes the meta-planner to maximize execution stability while minimizing computational overhead. This section elaborates on our dual-track methodology: (I) constructing a high-quality verifiable planning dataset, and (II) employing IGPO to establish robust architectural reasoning.

\vspace{-5pt}
\subsection{Data Construction}
To enable generative planning, we formulate the system design as a conditional code generation task. To bridge the lack of architectural priors in standard LLMs, we propose a \textit{Bootstrap-and-Filter} pipeline within \ourframework that transforms the search for optimal plans into a high-quality supervised dataset. This process involves four stages:

\vspace{-10pt}
\paragraph{Phase 1: Standardization via Unified Tool Interface.}
First, we utilize the modular nature of \ourframework to deconstruct the functional primitives of existing representative planning systems, specifically the 7 paradigms listed in Table~\ref{tab:plan_system_refined}. We decompose their discrete mechanisms into standardized tools. These tools are encapsulated within our unified framework, creating a shared \textit{Plan Space} where different topological structures can be expressed using a consistent code interface. 

\vspace{-8pt}
\paragraph{Phase 2: Evolutionary Sampling.}
With the standardized tools ready, we employ an evolutionary strategy to generate diverse planning candidates. For each query $Q_i$, we construct a specialized input context $\mathcal{C}_i$ consisting of:
\begin{itemize}[noitemsep, topsep=0pt, parsep=0pt, partopsep=0pt, leftmargin=*]
    \item The specific user query $Q_i$.
    \item The system prompt defining the Meta-Planner's role.
    \item Detailed documentation of the available Meta-Tools.
    \item A randomly sampled subset of 3 static planning samples $\{P_{ref}^1, P_{ref}^2, P_{ref}^3\}$ from our standardized pool, serving as structural references to guide the architectural design.
\end{itemize}
The model is tasked with synthesizing a unique, query-specific plan $P_{gen}$ by integrating or modifying these patterns to best suit $Q_i$. This process encourages the model to adapt the structural logic to the specific task requirements, rather than simply replicating existing templates.

\vspace{-5pt}
\paragraph{Phase 3: Execution-Based Verification.}
We validate each synthesized plan $P_{gen}$ by executing it within the PlanFactory runtime to generate a trajectory $\tau$ and final answer $A_{final}$. We apply a strict Execution-as-Judge filter: $P_{gen}$ is retained into the dataset if and only if $A_{final}$ matches the ground truth. This mechanism effectively purges hallucinated or unsound architectures, ensuring the Meta-Planner learns exclusively from successful design patterns.

\vspace{-5pt}
\paragraph{Phase 4: Preference Construction for SFT and IGPO.}
Finally, we format the validated execution trajectories into training supervision. To instill both correctness and efficiency into the Meta-Planner, we employ a dual-track alignment strategy, that separates fundamental capability learning from preference-based refinement:

    \textbf{SFT Data Construction:} During SFT, we adopt a strict outcome-supervised filtering protocol. We iterate through the generated plan candidates and retain only those pairs $(\mathcal{C}_i, P_{gen})$ that successfully execute. By grounding the target plan $P_{gen}$ on the reference-augmented context $\mathcal{C}_i$, we ensure that the base model learns to synthesize valid, executable architectures from the provided structural inspirations.

    \textbf{IGPO Data Construction:} To further align the model with high-quality planning logic via process supervision, we construct preference pairs $(P_{win}, P_{lose})$ for IGPO. We process the sampling results in pairs and determine the winner using a hierarchical criterion:
    
\begin{itemize}[noitemsep, topsep=0pt, parsep=0pt, partopsep=0pt, leftmargin=*]
        \item \textbf{Correctness First:} Correctness is the prerequisite. If one plan succeeds and the other fails, the successful plan is strictly preferred ($P_{win} \succ P_{lose}$).
        \item \textbf{Noise Filtering:} Pairs where both failed are discarded.
        \item \textbf{Efficiency as Tie-Breaker:} In ``expert scenarios'' where both candidates yield correct answers, we introduce a novel metric, \textbf{Cognitive Impedance} ($\mathcal{I}$), to resolve the tie. We define $\mathcal{I}$ as a compound cost function:
        \begin{equation} 
        \label{form:cognitive impedance}
        \small
        \mathcal{I}(\tau) = C_{tot} \cdot \exp\Big( \lambda_1 N_{fail} + \lambda_2 (1 - S_{stab}) + \lambda_3 \frac{C_{plan}}{C_{exec}} \Big) 
        \end{equation}
        \noindent where $C_{tot}$ is the total cost, $N_{fail}$ counts errors, and $S_{stab}$ quantifies execution smoothness. Crucially, the ratio of planning cost ($C_{plan}$) to execution cost ($C_{exec}$) acts as a bureaucracy penalty, ensuring planning effort does not outweigh execution.
    \end{itemize}

Formally, this pipeline yields two corpora: $\mathcal{D}_{SFT} = \{(\mathcal{C}_i, P_{gen}) \mid \text{Correct}(P_{gen}) \}$ for structural competence, and $\mathcal{D}_{IGPO} = \{(\mathcal{C}_i, P_{win}, P_{lose}) \mid P_{win} \succ P_{lose} \}$ for efficiency alignment.

\subsection{\ourmodel: Training Meta-Planner}

This section details the training methodology for \ourmodel. We optimize the Meta-Planner $\pi_{\theta}$ to synthesize planning configurations that maximize downstream agent performance. We adopt a two-stage curriculum: SFT establishes structural competence, followed by IGPO to align the planner with execution efficiency.

\subsubsection{Stage 1: Structural Competence via SFT}
We first instill the fundamental capabilities of code generation and architectural reasoning into the Meta-Planner. Leveraging $\mathcal{D}_{SFT}$, we treat the verified pairs $(\mathcal{C}, P{gen})$ as expert demonstrations. We optimize $\pi_{\theta}$ using the standard next-token prediction objective by minimizing the negative log-likelihood of the target sequence. This supervised training serves as a crucial warm-start phase, ensuring that the model acquires the necessary syntactic rules and API constraints. Consequently, it learns to synthesize valid instances of $P$ that are structurally grounded in the context $\mathcal{C}$, providing a stable initialization for subsequent alignment.

\subsubsection{Stage 2: Impedance-Guided Preference Alignment}

\setlength{\abovedisplayskip}{3pt}
\setlength{\belowdisplayskip}{3pt}

While SFT ensures syntactic viability, it does not guarantee execution efficiency. The subspace of functionally correct plans is vast, yet the subset of optimal configurations—those that minimize resource consumption while maximizing success—is sparse. To transition from static correctness to dynamic optimality, we formulate planning generation as a meta-level optimization problem.

Let $P \in \mathcal{P}$ denote an executable plan configuration. The Meta-Planner searches the plan space for an optimal configuration $P^*$ that maximizes the expected return, balancing task success against operational costs:
\begin{equation}
P^* = \arg\max_{P \in \mathcal{P}} \mathbb{E}_{\tau \sim \mathcal{M}(P)} [R(\tau) - \lambda \mathcal{I}(\tau)]
\end{equation}
where $R(\tau)$ is the binary success reward and $\mathcal{I}(\tau)$ represents the cognitive impedance. To solve this, we employ our IGPO method.

\vspace{-6pt}
\begin{table}[t!]
    \centering
    \caption{Detailed statistics of the constructed datasets. We operate in a long-context regime, where the input $L_{Context}$ ($\sim$13k tokens) is a composite sequence comprising the system prompt, tool definitions, retrieved structural examples, and the specific user query.}
    \label{tab:dataset_detailed}
    \begingroup 
    \fontfamily{ptm}\selectfont 
    \begin{tabular}{l|c|c|c|c}
        \toprule
        \textbf{Dataset Stage} & \textbf{Samples} & \textbf{Input} ($L_{Context}$) & \textbf{Reasoning} ($L_{CoT}$) & \textbf{Code} ($L_{Code}$) \\
        \midrule
        \textbf{Stage 1: SFT} & 3360 & $\sim$ 13,199  & $\sim$ 423  & $\sim$ 1,642   \\
        \textbf{Stage 2: IGPO} & 2000 & $\sim$ 13,168  & $\sim$ 497  & $\sim$ 1,636  \\
        \bottomrule
    \end{tabular}
    \endgroup
    \vspace{-2mm} 
\end{table}

\vspace{-5pt}
\paragraph{Impedance-Contrastive Rejection Sampling.}
Unlike standard preference collection which often relies on subjective human ranking, our framework constructs preference pairs based on objective execution metrics. The data curation process functions as a rejection sampling mechanism designed to distill efficiency signals from stochastic exploration:
\begin{itemize}[itemsep=3pt, parsep=0pt, topsep=2pt, leftmargin=*]
    \item \textbf{Exploratory Synthesis:} Given a context $\mathcal{C}$, the current policy $\pi_{\theta}$ samples $K$ candidate plans $\{\phi_1, \dots, \phi_K\}$, instantiating varied transition dynamics for the Agent System.
    
    \item \textbf{Execution \& Evaluation:} The Agent System executes these plans to generate trajectories $\tau_i$. Each trajectory is evaluated using the composite impedance metric $\mathcal{I}(\tau_i)$, aggregating token consumption, temporal latency, and runtime errors.
    
    \item \textbf{Contrastive Pair Construction:} We construct the preference dataset $\mathcal{D}_{IGPO}$ by selecting pairs $(\phi_{win}, \phi_{lose})$. To ensure functional validity, we enforce $R(\tau_{win})=1$. A pair is selected only if there exists a significant impedance gap $\mathcal{I}(\tau_{lose}) - \mathcal{I}(\tau_{win}) > \delta$, ensuring the optimization is driven by high-confidence efficiency signals.
\end{itemize}

\begin{table*}[!t]
\centering
\caption{Performance of various agent frameworks on the WebWalerQA, xBench-Ds, TaskCraft, and GAIA benchmarks. For each column, the best and second-best pass@1 scores are highlighted in bold and underlined respectively.}
\vspace{-0.5em}
\label{tab:main-table-1}
\setlength{\tabcolsep}{6pt}
\resizebox{\textwidth}{!}{
\begin{tabular}{ll|ccccccc}
\toprule
\midrule
\multirow{2}{*}{\textbf{Framework}} & \multirow{2}{*}{\textbf{Model Family}} & \multirow{2}{*}{\textbf{\makecell{WebWalker\\QA}}} & \multirow{2}{*}{\textbf{\makecell{xBench\\-DS}}} & \multirow{2}{*}{\textbf{\makecell{Task\\Craft}}} & \multicolumn{4}{c}{\textbf{GAIA}} \\
\cmidrule(lr){6-9}
& & & & & \textbf{Avg.} & \textbf{Level 1} & \textbf{Level 2} & \textbf{Level 3} \\
\Xhline{0.5pt}
OWL Workforce \hlblock{pass@3} & GPT-4o+o3-mini & 57.64 & 55.0 & 58.33  & 60.61 & 81.14 & 58.14 & 26.92 \\
OWL RP \hlblock{pass@3} & GPT-4o+o3-mini & - & - & - & 58.18 & 81.14 & 54.65 & 23.08 \\
TapeAgents & Claude 3.7 etc. & - &-  &-  & 55.76 & 71.70 & 53.49 & 30.77 \\
AutoAgent & Claude 3.5 etc. & - & - & - & 55.15 & 71.70 & 53.40 & 26.92 \\
Smolagents  & GPT-4.1 & - & - & - & 55.15 & 67.92 & 53.49 & 34.62 \\
Smolagents  & GPT-5-mini & 58.82 & 51.0 & 64.00 & 55.75 & 69.81 & 54.65 & 30.77 \\
Magnetic-1 & OpenAI o1 etc. & - & - & - & 46.06 & 56.60 & 46.51 & 23.08 \\
Cognitive Kernel-Pro  & Claude-3.7 etc. & 60.64 &  56.0 &  66.00  & 60.00 & 79.25 & 56.98 & 30.77 \\
Cognitive Kernel-Pro \hlblock{pass@3} & Claude-3.7 etc. & - & - & - & 75.15 & 84.91 & 73.26 & 61.54 \\
OAgents & Claude-3.7 etc. & 58.23 & 47.0 & - & 66.67 & 77.36 & 66.28 & \cellcolor{orange!20}\uline{46.15} \\
Agent KB  & GPT-4.1 &  60.59 & 48.0 &61.67  & 61.21 & 79.25 & 58.14 & 34.62 \\
Agent KB \hlblock{pass@2} & GPT-4.1 &  68.82 &  58.0 & 72.67 & 67.27 & 83.02 & 67.44 & 34.62 \\
Agent KB \hlblock{pass@3} & GPT-4.1 &  73.53 & 68.0 &  75.33 & 73.94 & 84.91 & 73.26 & 53.85 \\
Flash-Searcher   & GPT-5-mini & \cellcolor{orange!20}\uline{71.18} & 69.0 & 69.67 & 69.09 & 79.25 & \cellcolor{orange!20}\uline{69.77} & \cellcolor{orange!20}\uline{46.15} \\
Flash-Searcher  & Kimi K2 & 52.35 & 66.0 & 58.00 & 52.12 & 58.49 & 52.33 & 34.62 \\
Flash-Searcher  & DeepSeek V3.2 & 69.41 & 68.0 & 69.33 & 60.61 & 79.25 & 53.49 & \cellcolor{orange!20}\uline{46.15} \\

\midrule
\rowcolor[HTML]{EFEFEF} \ourmethod+ Smolagents & GPT-5-Mini & \cellcolor{orange!40}\textbf{73.53} & \cellcolor{orange!40}\textbf{75.0} & \cellcolor{orange!40}\textbf{72.67} &  \cellcolor{orange!40}\textbf{72.12} & \cellcolor{orange!20}\uline{81.14} & \cellcolor{orange!40}\textbf{72.09} & \cellcolor{orange!20}\uline{46.15} \\
\rowcolor[HTML]{EFEFEF} \ourmethod+ Smolagents & Kimi K2 & 64.71 & 71.0 & 69.33 & 60.00 & 73.58 & 55.81 & \cellcolor{orange!20}\uline{46.15} \\
\rowcolor[HTML]{EFEFEF} \ourmethod+ Smolagents & DeepSeek V3.2 & 70.59 & \cellcolor{orange!20}\uline{74.0} & \cellcolor{orange!20}\uline{71.33} & \cellcolor{orange!20}\uline{70.91} & \cellcolor{orange!40}\textbf{84.91} & 67.44 & \cellcolor{orange!40}\textbf{53.85} \\
\midrule
\bottomrule
\end{tabular}}
\vspace{-1em}
\end{table*}

\vspace{-8pt} 
\paragraph{Implicit Reward Alignment.}
We posit that the optimal policy $\pi^*$ should assign probability mass to a configuration $\phi$ inversely proportional to its impedance, subject to a KL-divergence constraint that prevents deviation from the reference distribution. Defining the implicit reward as $r(\phi) = - \mathbb{E}[\mathcal{I}(\tau)]$ for successful trajectories, the optimal policy follows the Boltzmann distribution:
\begin{equation}
\small
\pi^*(\phi \mid \mathcal{C}) \propto \pi_{ref}(\phi \mid \mathcal{C}) \cdot \exp \left( \frac{1}{\beta} r(\phi) \right)
\end{equation}
This formulation allows us to bypass training an explicit reward model. Following the DPO derivation, the implicit reward $r_{\theta}(\phi)$ can be re-parameterized by the log-ratio of the policy likelihoods:
\begin{equation}
\small
r_{\theta}(\phi) = \beta \log \frac{\pi_{\theta}(\phi \mid \mathcal{C})}{\pi_{ref}(\phi \mid \mathcal{C})}
\end{equation}
The final IGPO loss function maximizes the margin between efficient and inefficient architectures by minimizing:
\begin{equation}
\mathcal{L}_{IGPO}(\theta) = - \mathbb{E}_{(\phi_{w}, \phi_{l}) \sim \mathcal{D}_{IGPO}} \Big[ \log \sigma \big( r_{\theta}(\phi_{w}) - r_{\theta}(\phi_{l}) \big) \Big]
\end{equation}
This approach directly aligns the Meta-Planner with the execution environment, teaching it to architect systems that minimize cognitive impedance while maintaining functional correctness.

\section{Experiments}

\subsection{Experiment Setup}

\paragraph{Training.}
To equip our model with robust planning capabilities, we construct a high-quality composite dataset sourced from diverse domains. Our training corpus aggregates samples from TaskCraft~\citep{shi2025taskcraftautomatedgenerationagentic}, MoNaCo~\citep{wolfson2026monaco}, WebWalkerQA~\citep{wu2025webwalkerbenchmarkingllmsweb}, and DeepSearchQA~\citep{kaggleDeepSearchQA}.The data construction pipeline leverages a teacher-student paradigm, utilizing Gemini-3-Flash as the expert planner to generate high-level reasoning traces, and DeepSeek V3.2 as the executor to verify actionable outcomes.The final curated dataset detail is shown in Table \ref{tab:dataset_detailed}.
We employ Qwen3-14B~\citep{yang2025qwen3technicalreport} as our backbone model. 

\vspace{-12pt}
\paragraph{Testing \& Baselines.}
To rigorously evaluate the model's ability to handle diverse and multimodal queries, we employ a comprehensive evaluation suite. Our benchmarks include the complete GAIA~\citep{mialon2023gaia} and XBench-DS~\citep{chen2025xbenchtrackingagentsproductivity}. Additionally, we construct specific test splits from TaskCraft~\citep{shi2025taskcraftautomatedgenerationagentic} and WebWalkerQA~\citep{wu2025webwalkerbenchmarkingllmsweb}. Crucially, the test samples from these datasets are distinct and non-overlapping with the training splits to prevent data leakage.
For fair comparison during inference, the underlying LLMs driving the agents include DeepSeek V3.2~\citep{deepseekai2025deepseekv32pushingfrontieropen}, Kimi-K2~\citep{kimiteam2025kimik2openagentic}, and GPT-5-mini~\citep{openaiIntroducingGPT52}. We utilize Gemini-3-Flash~\citep{comanici2025gemini} as the judge model to provide unbiased evaluation of agent trajectories.
To validate efficacy, we benchmark \ourmodel~against a wide spectrum of state-of-the-art systems. Please refer to Table~\ref{tab:main-table-1} for the detailed list of all baselines compared.

\begin{table*}[!t]
\centering
\caption{Comprehensive comparison of execution performance across different agent frameworks. The framework achieving the highest accuracy on each benchmark is highlighted in bold.}
\label{tab:efficiency_comparison}
\vspace{-0.5em}
\setlength{\tabcolsep}{3.5pt} 
\resizebox{\textwidth}{!}{
\begin{tabular}{ll|ccccccc>{\columncolor[HTML]{EFEFEF}}c}
\toprule
\textbf{Benchmark} & \textbf{Metric} & \textbf{Co-Sight} & \textbf{FlowSearch} & \textbf{Flash-Searcher} & \textbf{AgentOrchestra} & \textbf{OAgents} & \textbf{JoyAgent} & \textbf{OWL} & \textbf{\ourmethod} \\
\midrule

\multirow{5}{*}{\textbf{\makecell{WebWalker-QA}}} 
 & Accuracy (\%) & 16.67 & 30.00 & 60.00 & 46.67 & 33.33 & \cellcolor{orange!20}{63.33} & 53.33 & \cellcolor{orange!40}\textbf{70.00}\\
 & Avg Cost (\$) & 0.0013 & 0.0053 & 0.0134 & 0.0112 & 0.0236 & 0.0028 & 0.0062 & 0.0167 \\
 & Avg Time (s) & 190.52 & 94.79 & 164.78 & 137.69 & 150.74 & 212.83 & 127.63 & 216.59\\
 & Avg Step & 2.1 & 4.0 & 5.3 & 6.5 & 7.2 & 4.0 & 3.8 & 7.7\\
\cmidrule(lr){1-10}

\multirow{5}{*}{\textbf{\makecell{DeepSearch-QA}}} 
 & Accuracy (\%) & 4.00 & 16.00 & 22.00 & 20.00 & 28.00 & 28.00 & \cellcolor{orange!20}{30.00} &  \cellcolor{orange!40}\textbf{42.00} \\
 & Avg Cost (\$) & 0.0025 & 0.0109 & 0.0408 & 0.0263 & 0.0454 & 0.0034 & 0.0191 & 0.0495 \\
 & Avg Time (s) & 895.88 & 351.76 & 522.36 & 437.06 & 519.91 & 548.70 & 428.63 & 875.26 \\
 & Avg Step & 2.8 & 5.5 & 10.0 & 9.9 & 10.8 & 4.0 & 6.9 & 11.7\\
\cmidrule(lr){1-10}

\multirow{5}{*}{\textbf{GAIA-level2 Text-only}} 
 & Accuracy (\%) & 17.14 & 25.71 & 25.71 & 14.29 & 15.71 & \cellcolor{orange!20}{30.00} & 24.29 & \cellcolor{orange!40}\textbf{57.14} \\
 & Avg Cost (\$) & 0.0018 & 0.0069 & 0.0255 & 0.0149 & 0.0317 & 0.0027 & 0.0130 & 0.0282\\
 & Avg Time (s) & 250.23 & 159.14 & 305.67 & 222.75 & 292.12 & 304.38 & 299.78 & 323.65\\
 & Avg Step & 2.6 & 4.6 & 8.0 & 7.7 & 8.7 & 4.1 & 6.2 & 9.1\\
 
\bottomrule
\end{tabular}
}
\vspace{-0em}
\end{table*}

\vspace{-4pt}
\subsection{Main Results}
\label{sec:main_results}
\textbf{Substantial Performance Enhancement over Baselines.} As presented in Table~\ref{tab:main-table-1}, integrating \ourmethod with the Smolagents framework yields significant performance gains across all evaluated benchmarks. On the comprehensive GAIA benchmark, our approach using GPT-5-Mini achieves an average score of 72.12\%, marking a remarkable absolute improvement of 16.37\% over the vanilla Smolagents baseline. Furthermore, our method outperforms specialized frameworks operating with the same backbone; for instance, it surpasses Flash-Searcher on GAIA Avg and demonstrates superior versatility on domain-specific benchmarks like WebWalkerQA and xBench-DS. These results empirically validate that the autonomous synthesis of task-specific planning architectures offers greater adaptability than static graph-based priors.

\vspace{-5pt}
\textbf{Consistent Gains across Diverse Backbones.} 
The scalability of \ourmethod is evidenced by its consistent improvements across diverse execution backbones, including GPT-5-Mini, DeepSeek V3.2 and Kimi K2. Notably, when equipped with the DeepSeek V3.2, our framework achieves a GAIA average of 70.91\%, significantly outperforming the Flash-Searcher implementation using the same model by over 10 percentage points. This consistency suggests that the meta-planner acquires transferable architectural reasoning capabilities that function independently of the execution model's internal knowledge, effectively acting as a general-purpose performance booster for agentic systems.

\vspace{-5pt}
\textbf{Complex Reasoning with Open-Source Frameworks.} The advantages of \ourmethod are particularly pronounced in high-complexity scenarios requiring long-horizon reasoning. On GAIA Level 3, the most challenging subset, our framework driven by DeepSeek V3.2 attains a success rate of 53.85\%. This performance not only surpasses the standard Agent KB using the more powerful GPT-4.1 but also matches the performance of Agent KB with pass@3 voting . This finding highlights a critical insight: with optimal dynamic planning topology, cost-effective open-weights models can rival or exceed the capabilities of resource-intensive proprietary models in complex problem-solving.

\begin{figure}[!t] 
    \centering
    \includegraphics[width=0.8\linewidth]{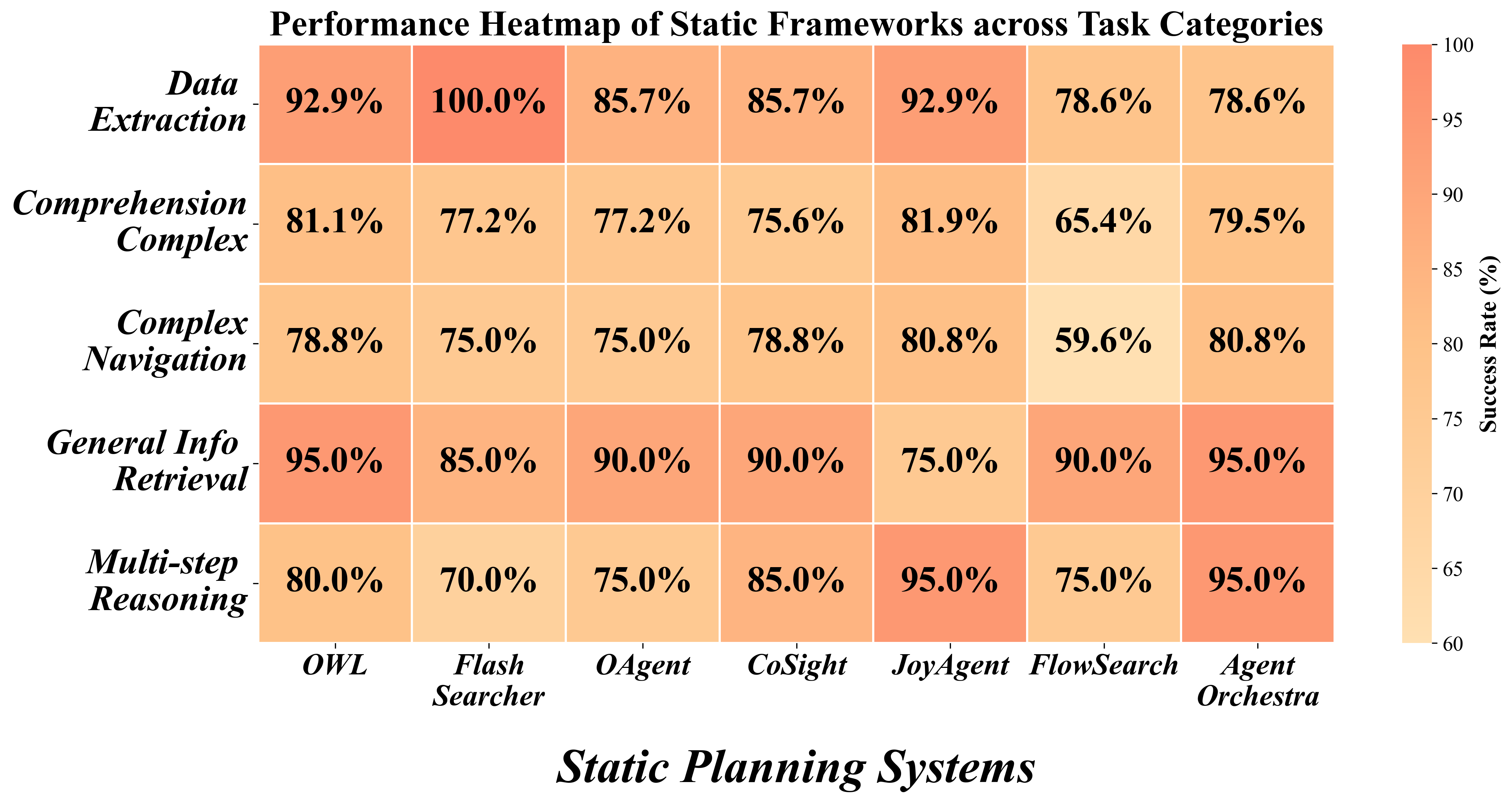} 
    \vspace{-0.1em} 
    
    \caption{Task-Dependent Performance Variability.}
    \label{fig:heatmap_system}
    \vspace{-0.1em} 
\end{figure}

\vspace{-1pt}
\subsection{Structural Specialization} 
We first investigate the performance variability of fixed planning architectures across diverse task typologies, leveraging the GPT-5-mini~\citep{openaiIntroducingGPT52} to evaluate a multi-category benchmark extracted from TaskCraft~\citep{shi2025taskcraftautomatedgenerationagentic}. As visualized in Figure~\ref{fig:heatmap_system}, distinct planning priors exhibit strong inductive biases suitable for specific domains but lack universality. For instance, centralized systems trade data-handling capacity for reasoning depth, whereas DAG topologies prioritize extraction speed over logical coherence. This heterogeneity highlights a critical limitation that rigid topologies cannot optimally address the structural diversity of open-ended queries. This empirical evidence validates the core premise of \ourmethod: by dynamically synthesizing architectures that integrate the complementary strengths of diverse planning paradigms, our meta-planner achieves cross-domain robustness that no single static framework can match.

\vspace{-1pt}
\subsection{Inference Efficiency} 
Beyond task adaptability, we evaluate whether the performance gains of \ourmethod{} come at the expense of excessive computational overhead. Table~\ref{tab:efficiency_comparison} details the execution metrics on three benchmarks using the Kimi-K2~\citep{kimiteam2025kimik2openagentic} backbone. \ourmethod{} consistently achieves dominant accuracy, surpassing the best static baseline by substantial margins (e.g., +10.0\% on WebWalker-QA, +14.0\% on DeepSearch-QA). Crucially, this performance does not incur a proportional spike in resource consumption, \ourmethod{} demonstrates superior Pareto optimality: it maintains comparable costs and latency to sophisticated baselines while delivering significantly higher success rates. This indicates that the meta-planner effectively minimizes cognitive impedance, avoiding the redundant loops of inefficient planners and the premature failures of overly simple ones.

\begin{figure}[t] 
    \centering
    \includegraphics[width=\linewidth]{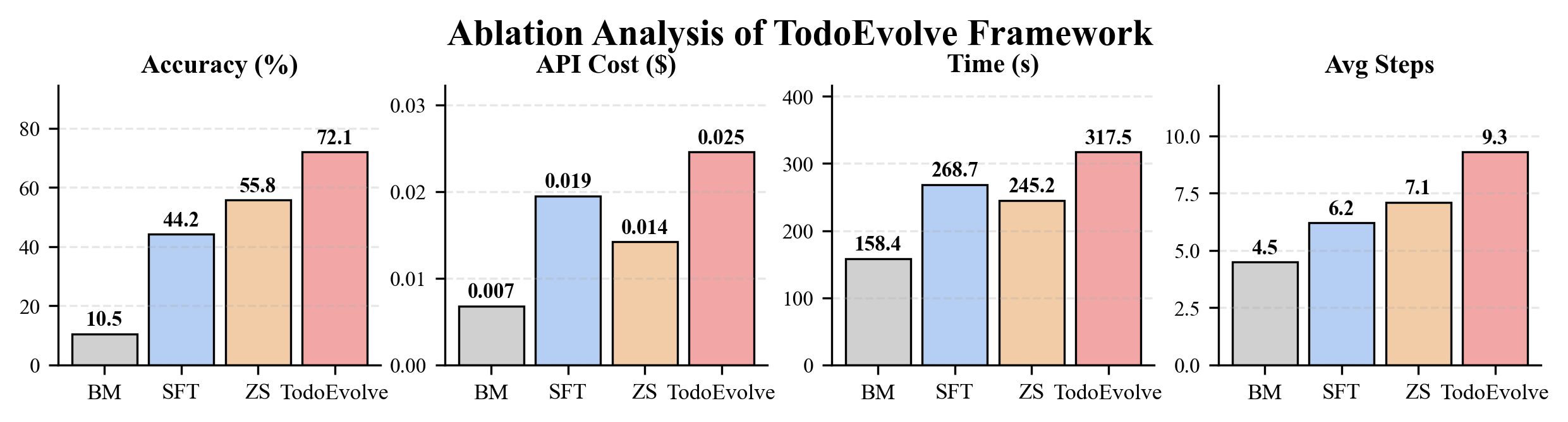} 
    \vspace{-1.4em} 
   
    \caption{Ablation Analysis on GAIA Level 2. We compare the following variants, BS (Base Model), SFT (SFT-Only), ZS (Zero-Shot) and \ourmethod.}
    \label{fig:ablation_study}
    \vspace{-0em} 
\end{figure}

\begin{figure*}[t]
  \centering
  \includegraphics[width=1\textwidth]{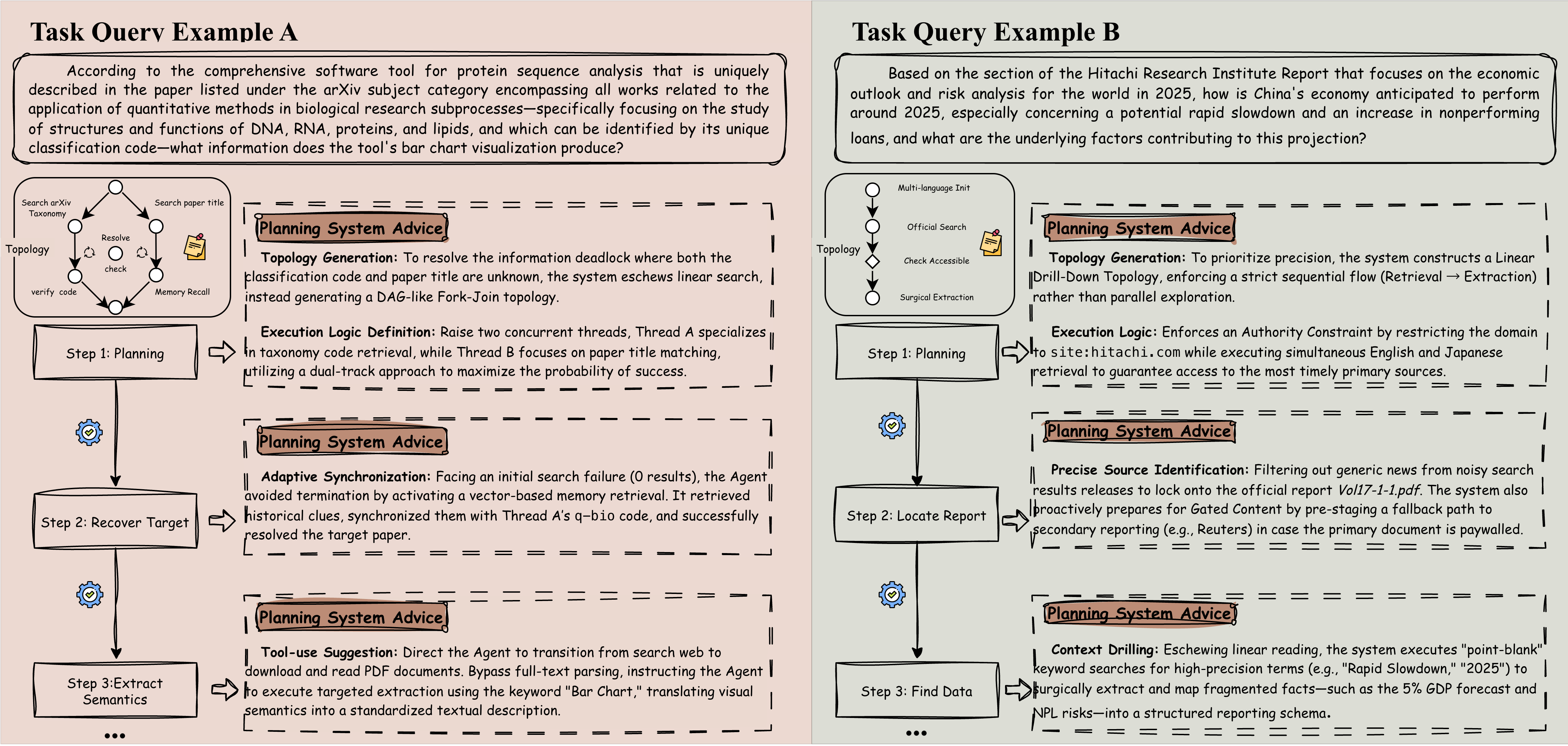}
  \vspace{-1pt} 
  \caption{Evolved planning architectures in real-world instantiation. The system provides adaptive, state-aware structural scaffolding that spans from macro-topology initialization to granular adaptation and navigation during the execution stage, effectively steering the agent toward robust and resilient inference.}
  \label{fig:case_study}
  \vspace{-2pt} 
\end{figure*}

\subsection{Ablation Study} 
To dissect the efficacy of our training components, we conduct an ablation study on the GAIA Level 2 validation set, comparing four configurations: (1) Base Model, utilizing the unaligned Qwen3-14B to generate planning systems; (2) SFT-Only, fine-tuned exclusively on verified planning trajectories; (3) Zero-Shot, which incorporates our IGPO training but performs inference without few-shot examples; and (4) \ourmethod, the complete framework employing both training stages and reference-augmented inference. As illustrated in Figure~\ref{fig:ablation_study}, the Base Model fails to synthesize executable plans due to a lack of syntactic grounding, a capability established by SFT-Only. Notably, the {Zero-Shot setting not only improves accuracy to 55.8\% but also reduces API costs relative to SFT-Only, confirming that IGPO effectively optimizes execution efficiency. Finally, \ourmethod achieves a peak accuracy of 72.1\%; the concomitant increase in steps and cost reflects the planner's enhanced capability to persist through and resolve complex, long-horizon tasks that simpler variants abandon.

\vspace{-1pt}
\subsection{Case Study}
\vspace{-0.3em}
To intuitively illustrate how \ourmethod facilitates complex reasoning, we present a qualitative analysis of a planning system synthesized during a real execution. As shown in Figure~\ref{fig:case_study}, unlike static, "one-size-fits-all" scaffolds, \ourmethod delivers a dynamic planning architecture that is adaptively tailored to the evolving task state.

We present a qualitative analysis of the planning system synthesized during real execution, as shown in Figure~\ref{fig:case_study}. The results illustrate that \ourmethod delivers a dynamic planning architecture that is adaptively tailored to the evolving task state. Specifically, the planner identifies the optimal computational shape for impedance reduction: it instantiates a high-breadth Fork-Join topology to break information deadlocks (\textit{Task A}), while conversely enforcing strict linear constraints to prune search-space noise for high-precision targets (\textit{Task B}). Notably, the system exhibits predictive resilience by anticipating access barriers—such as paywalled reports—and proactively staging fallback paths to secondary sources. Together, these mechanisms ensure the plan acts as a state-aware anchor, preventing reasoning drift and transforming passive generation into active, strategic solving.

We present more concrete visualizations of the planning systems designed by \ourmodel in \Cref{app:case}.

\vspace{-0.1em}
\section{Conclusion}
\vspace{-0.1em}
Traditional agentic planning relies on "one-size-fits-all" workflows, often proving rigid and suboptimal for diverse task demands. This paper aims to transform planning from manual engineering into an autonomous synthesis process, making architectural design as adaptive as the underlying model’s reasoning. To this end, we introduce \textbf{\ourmethod}, a meta-planning paradigm that navigates a unified design space, \textbf{\ourframework}, to dynamically configure task-specific topologies and strategies via IGPO. Our extensive evaluations across diverse benchmarks demonstrate that \ourmethod outperforms static baselines, achieving Pareto optimality between success rates and computational efficiency. By bridging the gap between internal reasoning and external architectural scaffolding, \ourmethod provides a blueprint for self-evolving agents capable of mastering open-ended, long-horizon complexities.

\clearpage
\section*{Contributions}

\begin{multicols}{2}
\textbf{Core Contributors}
\begin{itemize}
    \item Jiaxi Liu
    \item Yanzuo Jiang
\end{itemize}
\textbf{Project Lead}
\begin{itemize}
    \item Guibin Zhang
\end{itemize}
\textbf{Contributors}
\begin{itemize}
    \item Zihan Zhang
    \item Heng Chang
\end{itemize}

\textbf{Corresponding Authors}
\begin{itemize}
\item Zhenfei Yin
\item Qibing Ren
\item Junchi Yan
\end{itemize}


\end{multicols}


\clearpage
\bibliographystyle{apalike}
\bibliography{cite}

\appendix

\section{\ourframework Details}
\label{app:factory}

We detail the established planning system in \ourframework as follows:

\begin{itemize}
    \item \textbf{Co-Sight}
    
    Co-Sight establishes a cross-check net topology, specifically engineered to resolve epistemic discrepancies through mutual verification. The system is initialized via an inconsistency trigger, where the planning process is activated only upon detecting conflicting information or divergent perspectives among internal modules. Navigation is executed through conflict resolution, utilizing trustworthy reasoning with structured facts to systematically eliminate cognitive biases across the agent collective. For its adaptation mechanism, the framework employs meta-verification, conducting high-level assessments of the underlying verification logic to ensure the integrity of the process of building consensus.
    
    \item \textbf{AgentOrchestra}

    AgentOrchestra adheres to an orchestration hierarchy topology, establishing a structured command chain for multi-agent coordination. The system initiates through role definition, where functional identities are assigned to activate the environment. During this phase, a planning agent leverages its global perspective to decompose complex objectives into manageable sub-tasks. Navigation is facilitated via centralized routing, with the planning agent dispatching specific instructions to specialized sub-agents based on their designated roles. The framework’s adaptation is driven by environment feedback, where the system dynamically re-calibrates the plan by synthesizing execution data, aggregating feedback loops, and monitoring cumulative progress toward the final objective.

    \item \textbf{OAgents}

    OAgents employs a modular graph topology, representing the global objective as a web of decoupled yet interdependent modules. The framework initiates via SOP configuration, where the agent decomposes the primary task into sub-tasks interconnected by edges that define prerequisite dependencies. Navigation is driven by dynamic programming, which, at each discrete step, identifies and dispatches the set of candidate nodes whose dependencies have been fully satisfied. The system’s adaptation mechanism relies on critic-loop feedback for periodic refinement: every $N$ steps, intermediate results are cross-referenced against global constraints to verify alignment with the objective, triggering a re-sequencing of sub-tasks based on novel observations. Furthermore, trajectories from prior execution attempts are distilled into heuristic guidance and integrated into the planning module as soft constraints or behavioral preferences, dynamically biasing sub-task selection toward proven success paths.
    
    \item \textbf{JoyAgent} 
    
    JoyAgent utilizes a collective hierarchy topology, structuring its multi-agent system to balance global oversight with local flexibility. the system is initialized through hybrid planning, which implements a supervisor agent based on a plan-and-execute framework to maintain global coherence while concurrently deploying multiple single agents utilizing react to ensure step-level responsiveness. navigation is governed by joint deliberation, where outputs from the diverse agent pool are aggregated and processed through consensus voting to determine the optimal execution path. the framework’s adaptation is achieved through the intrinsic react loops of the individual agents, allowing for real-time adjustments based on localized feedback without compromising the overarching trajectory.
    
    \item \textbf{Flash-Searcher} 
    
    Upon receiving a request, Flash-Searcher decomposes the task into a parallel Directed Acyclic Graph (DAG), where nodes denote granular sub-tasks and edges represent their dependencies. The system instantiates this structure through dependency parsing, mapping out the prerequisite constraints to initialize the graph's nodes and edges. Navigation is governed by aggressive parallelization. A node is dispatched to a concurrent execution pool as soon as its predecessors are satisfied or when partial execution results provide sufficient auxiliary validation. To maintain system agility, the framework performs workflow pruning at defined step intervals, where it summarizes progress to excise resolved nodes and re-evaluates the dependencies of pending tasks, dynamically injecting new decomposition branches if environmental contingencies arise.
    
    \item \textbf{FlowSearch} 
    
    FlowSearch conceptualizes task resolution through a thought graph topology, representing the reasoning process as an evolving network of cognitive states. The framework employs flow construction for incremental instantiation; starting from the root task, a knowledge flow planner iteratively evaluates whether active nodes require further decomposition or supplemental context. This process generates descendant nodes that encapsulate sub-problems, intermediate reasoning steps, and required evidentiary grounding while concurrently establishing dependency edges to preserve logical consistency and structural integrity. Navigation is managed by a knowledge collector, which identifies and dispatches nodes that exhibit the highest execution readiness based on satisfied dependencies. The system’s adaptation is realized through dynamic expansion via a knowledge refiner, which leverages newly acquired insights to perform structural transformations on the flow. By synthesizing current knowledge contexts with execution states, the refiner dynamically executes atomic operations including the addition, deletion, or modification of nodes and edges to optimize the graph’s trajectory toward the goal.
    
    \item \textbf{OWL} 
    
    OWL adopts a dual hierarchy topology that formally segregates the strategic management layer from the tactical execution layer. Upon task arrival, the system undergoes planner decomposition, where a high-level planner analyzes task complexity against the latent capabilities of available worker nodes to instantiate a structured task list. Navigation is facilitated via dynamic dispatch, managed by a coordinator that evaluates real-time agent profiles to map specific sub-tasks to the most suitable worker nodes. The framework’s adaptation logic is driven by manager intervention triggered by decentralized failure detection: individual workers autonomously monitor their execution status, broadcasting failure signals to a dedicated task channel upon impasse. This channel acts as an observation primitive, prompting the planner to perform reactive re-planning and inject revised sub-tasks based on the contextual feedback from the failed execution.

\end{itemize}

\section{Datasets}

The five datasets used in this study are described as follows: (1) \textbf{GAIA} \citep{mialon2023gaia} consists of $165$ tasks, categorized into $53$ Level-1, $86$ Level-2, and $26$ Level-3 problems. (2) \textbf{WebWalkerQA} \citep{wu2025webwalkerbenchmarkingllmsweb} evaluates an agent’s capability in handling complex, multi-turn web interactions. It comprises $680$ real-world queries across four domains and spans over $1,373$ webpages. We sample a subset of $170$ queries for evaluation. (3) \textbf{xBench-DeepSearch (xBench-DS)} \citep{chen2025xbenchtrackingagentsproductivity} contains $100$ tasks assessing agentic planning, tool use, and reasoning. (4) \textbf{TaskCraft}\citep{shi2025taskcraftautomatedgenerationagentic} is a synthetic benchmark generated via an autonomous data pipeline, we collect $300$ queries as a valid subset.(5) \textbf{DeepSearchQA} \citep{kaggleDeepSearchQA} targets the long-horizon research capabilities of agents, we collect $50$ queries as a valid subset.

\section{Case Study}\label{app:case}

To provide a concrete and intuitive understanding of the planning architectures synthesized by \ourmethod, we visualize three representative systems generated for distinct query types, as shown in Figures~\ref{fig:case1} to~\ref{fig:case3}. These examples demonstrate how our meta-planner moves beyond static templates, dynamically tailoring the control flow—ranging from linear sequential logic to complex parallel graph structures—to match the specific cognitive impedance and dependency requirements of the task. By autonomously configuring the topology initialization, execution navigation, and adaptation triggers, \ourmethod ensures robust performance across varying levels of problem complexity.

\begin{figure*}[!h]
    \centering
    \includegraphics[width=0.5\textwidth]{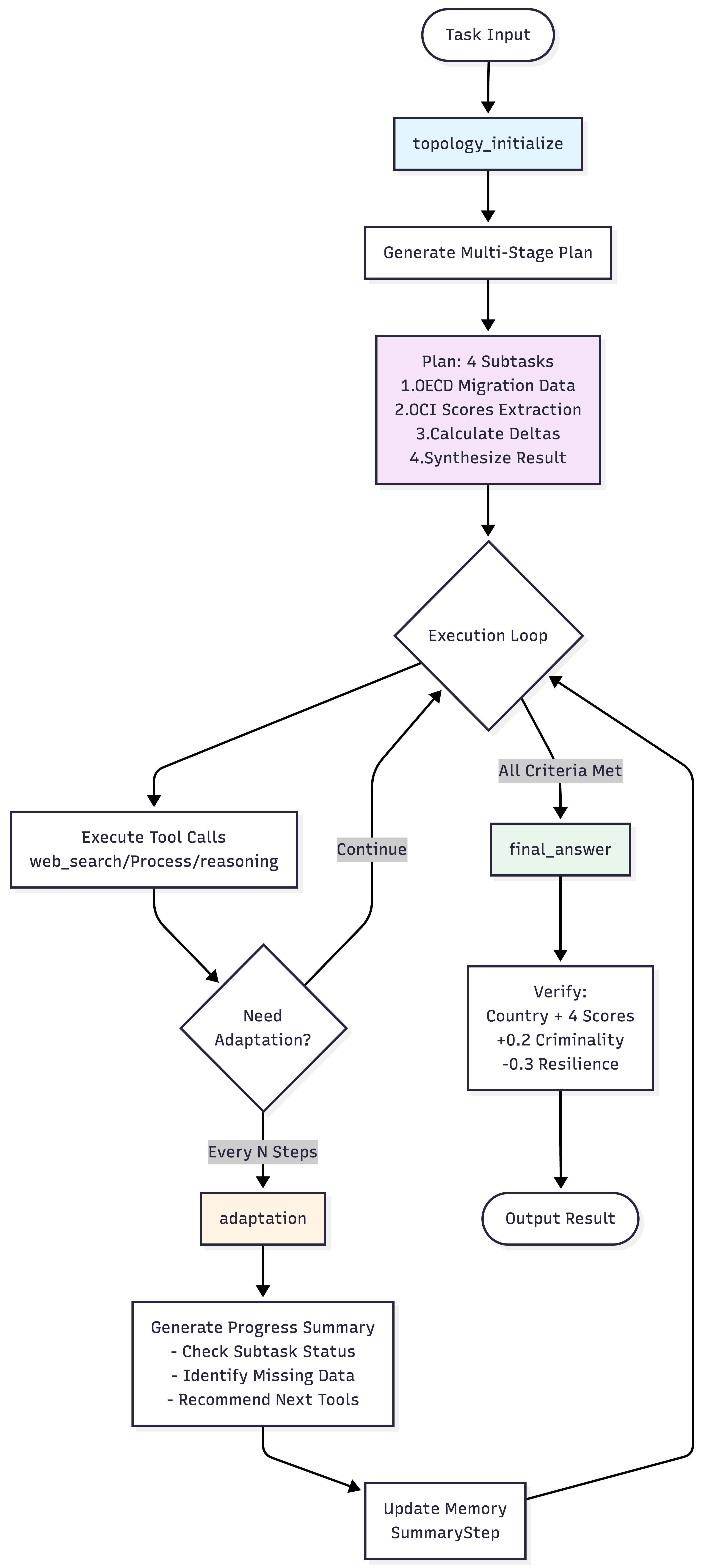}
    \caption{\textbf{Linear Sequential Planning for Multi-Criteria Filtering.} 
    For a query requiring strict multi-stage filtering and calculation (identifying countries based on migration thresholds followed by crime index analysis), \ourmethod instantiates a linear execution topology. The system prioritizes a sequential ``fetch-and-filter'' pipeline to manage data dependencies, incorporating a periodic adaptation trigger to validate intermediate retrieval results before proceeding to the final synthesis and verification stage. This structure minimizes branching overhead for tasks where step-wise logical progression is paramount.}
    \label{fig:case1}
\end{figure*}

\begin{figure*}[!h]
    \centering
    \includegraphics[width=0.5\textwidth]{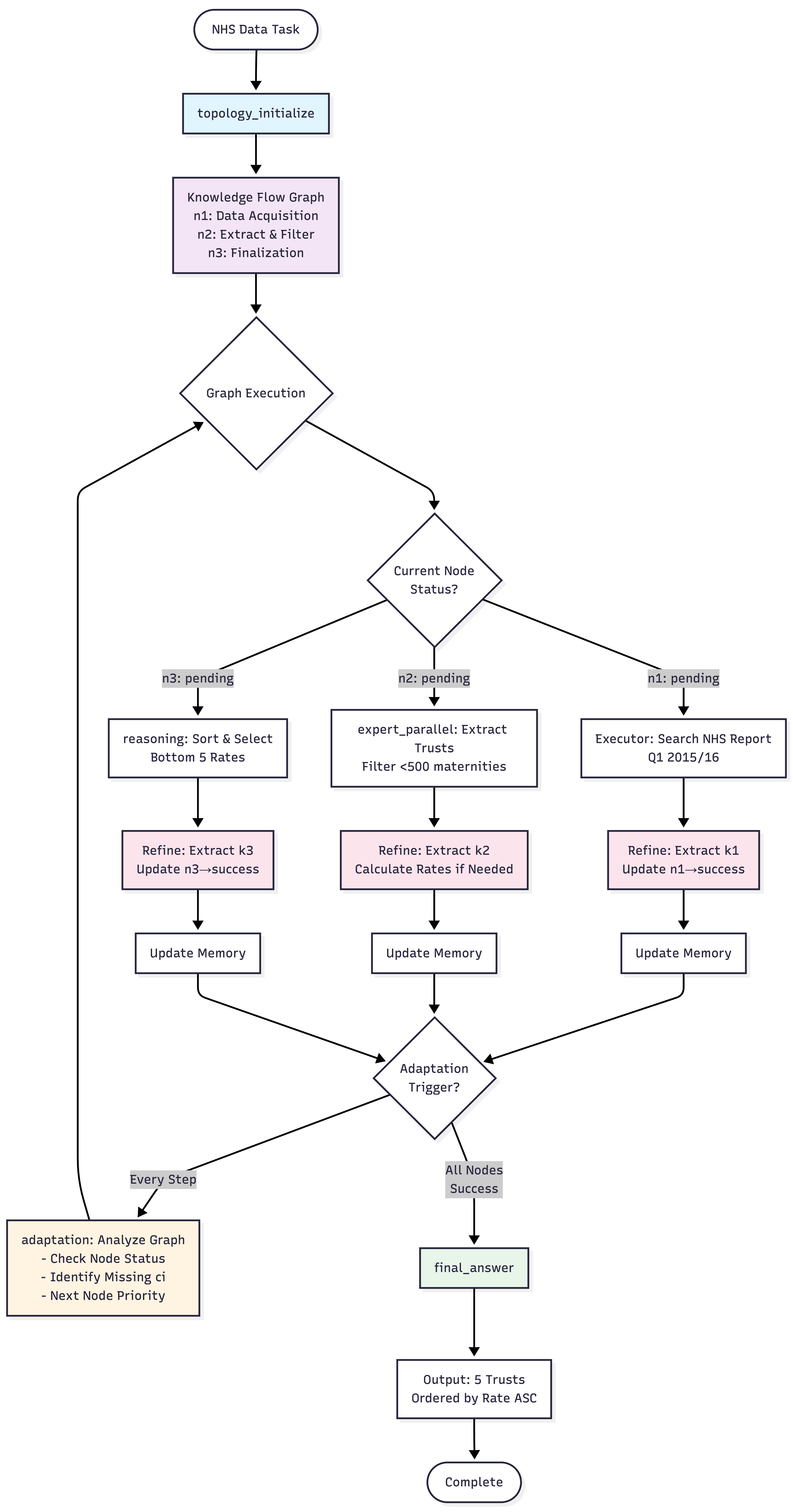}
    \caption{\textbf{State-Aware Graph Topology for Structured Data Extraction.} 
    Addressing a structured retrieval task involving sorting and ranking constraints, the meta-planner constructs a Knowledge Flow Graph. This topology decomposes the problem into granular nodes (acquisition, filtering, and finalization). The navigation strategy employs a state-aware routing mechanism that dynamically selects between parallel extraction or sequential reasoning based on the current node status ("pending" vs. "success"), allowing the system to efficiently prune the search space while adhering to numerical constraints.}
    \label{fig:case2}
\end{figure*}

\begin{figure*}[!h]
    \centering
    \includegraphics[width=0.7\textwidth]{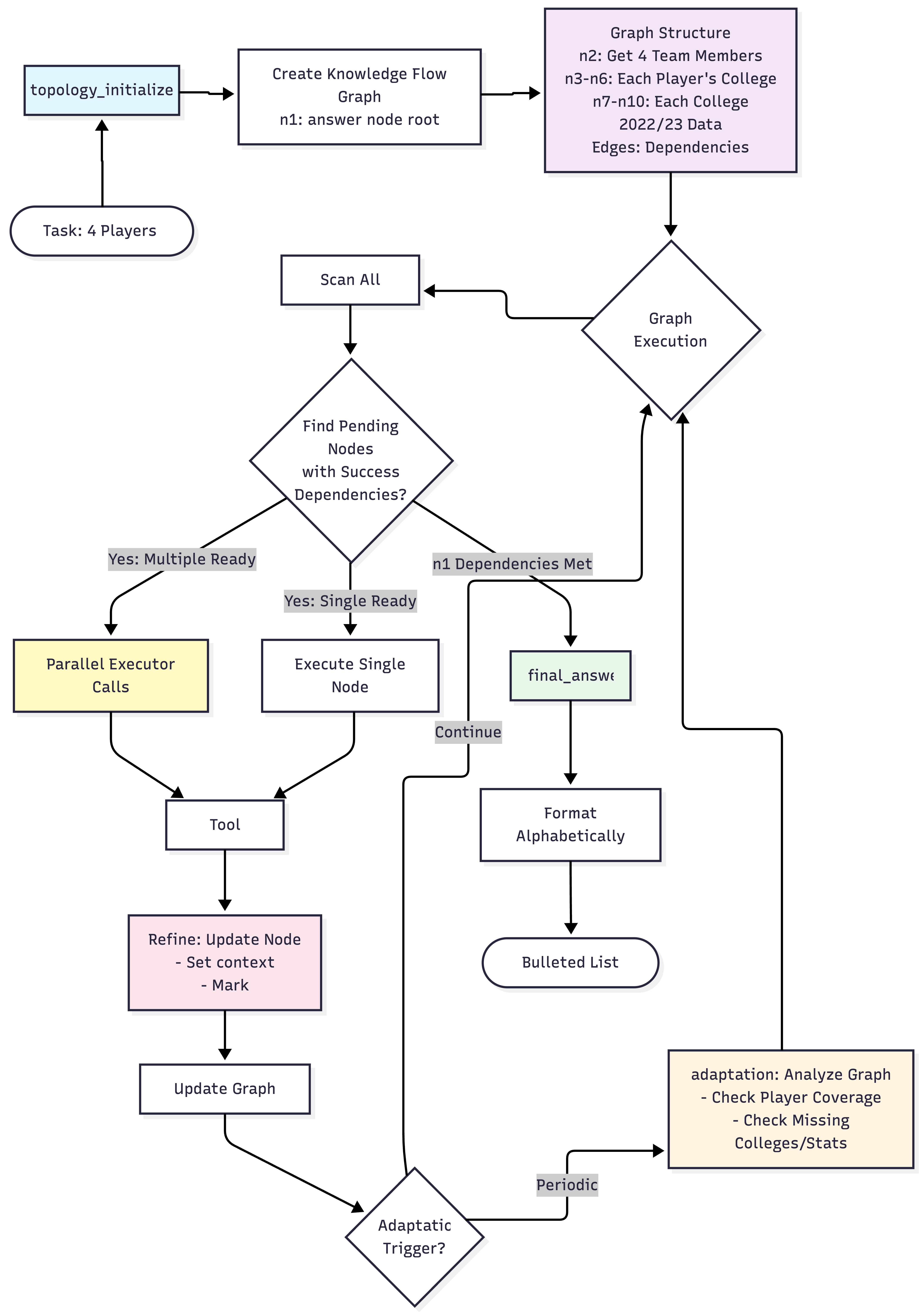}
    \caption{\textbf{High-Breadth Parallel Planning for Complex Entity Resolution.} 
    Faced with a complex entity resolution task requiring the retrieval of nested attributes for multiple subjects simultaneously, \ourmethod evolves a highly parallelized graph architecture. The system identifies independent sub-goals (e.g., retrieving data for different players concurrently) and activates a ``Parallel Executor'' module to minimize latency. The adaptation layer monitors the synchronization of these concurrent streams, ensuring that the graph topology is only updated and merged when specific dependency conditions are met.}
    \label{fig:case3}
\end{figure*}

\end{document}

%% file: common.tex
\usepackage{latexsym}
\usepackage[T1]{fontenc}
\usepackage[utf8]{inputenc}
\usepackage{microtype}
\usepackage{inconsolata}
\usepackage{graphicx}
\usepackage{hyperref}       
\usepackage{url}            
\usepackage{booktabs}       
\usepackage{amsfonts}       
\usepackage{nicefrac}       
\usepackage{stackengine}
\usepackage{microtype}      
\usepackage{colortbl}
\usepackage{xcolor}
\usepackage{amsmath}
\usepackage{amssymb}
\usepackage{amsthm}
\usepackage{mathrsfs}
\usepackage{pifont}
\usepackage{MnSymbol}
\usepackage{balance}
\usepackage{enumitem}
\usepackage{listings}
\usepackage{xcolor}
\usepackage{natbib}
\usepackage{multicol}

\AtBeginDocument{%
  \providecommand\BibTeX{{%
    \normalfont B\kern-0.5em{\scshape i\kern-0.25em b}\kern-0.8em\TeX}}}

\makeatletter
\DeclareRobustCommand\onedot{\futurelet\@let@token\@onedot}
\def\@onedot{\ifx\@let@token.\else.\null\fi}

\usepackage{setspace}
\usepackage{mathtools}

\usepackage{multirow,booktabs}
\usepackage{subcaption}

\newcommand{\owo}[1]{\textsc{OAgents}}

\definecolor{lightgreen}{RGB}{144, 238, 144} 
\definecolor{lightred}{RGB}{255, 105, 97}

\newtcolorbox{promptbox}[2][Prompt]{
colback=black!5!white,
arc=5pt, 
boxrule=0.5pt,
fonttitle=\bfseries,
title=#1, 
before upper={\small}, fontupper=\fontfamily{ptm}\selectfont,
colframe=#2, 
}
\definecolor{ogreen}{RGB}{34, 139, 34}

\definecolor{JDOrange}{RGB}{230, 126, 34} 
\definecolor{JDBurntOrange}{RGB}{211, 84, 0} 
\definecolor{JDLightOrange}{RGB}{255, 242, 230} 

\hypersetup{
  colorlinks=true,
  linkcolor=JDOrange,
  citecolor=JDBurntOrange,
  urlcolor=JDOrange,
  filecolor=JDOrange,
}

%% file: paper.bbl
\begin{thebibliography}{}

\bibitem[Andon, 2025]{andonlabsVendingBenchAndon}
Andon (2025).
\newblock {V}ending-{B}ench 2 | {A}ndon {L}abs --- andonlabs.com.
\newblock \url{https://andonlabs.com/evals/vending-bench-2}.
\newblock [Accessed 15-01-2026].

\bibitem[Backlund and Petersson, 2025]{backlund2025vendingbenchbenchmarklongtermcoherence}
Backlund, A. and Petersson, L. (2025).
\newblock Vending-bench: A benchmark for long-term coherence of autonomous agents.

\bibitem[Besta et~al., 2023]{got}
Besta, M., Blach, N., Kubicek, A., Gerstenberger, R., Gianinazzi, L., Gajda, J., Lehmann, T., Podstawski, M., Niewiadomski, H., Nyczyk, P., and Hoefler, T. (2023).
\newblock Graph of thoughts: Solving elaborate problems with large language models.

\bibitem[Cao et~al., 2025]{cao2025largelanguagemodelsplanning}
Cao, P., Men, T., Liu, W., Zhang, J., Li, X., Lin, X., Sui, D., Cao, Y., Liu, K., and Zhao, J. (2025).
\newblock Large language models for planning: A comprehensive and systematic survey.

\bibitem[Chen et~al., 2025]{chen2025xbenchtrackingagentsproductivity}
Chen, K., Ren, Y., Liu, Y., Hu, X., Tian, H., Xie, T., Liu, F., Zhang, H., Liu, H., Gong, Y., Sun, C., Hou, H., Yang, H., Pan, J., Lou, J., Mao, J., Liu, J., Li, J., Liu, K., Liu, K., Wang, R., Li, R., Niu, T., Zhang, W., Yan, W., Wang, X., Zhang, Y., Hung, Y.-H., Jiang, Y., Liu, Z., Yin, Z., Ma, Z., and Mo, Z. (2025).
\newblock xbench: Tracking agents productivity scaling with profession-aligned real-world evaluations.

\bibitem[Comanici et~al., 2025]{comanici2025gemini}
Comanici, G., Bieber, E., Schaekermann, M., Pasupat, I., Sachdeva, N., Dhillon, I., Blistein, M., Ram, O., Zhang, D., Rosen, E., et~al. (2025).
\newblock Gemini 2.5: Pushing the frontier with advanced reasoning, multimodality, long context, and next generation agentic capabilities.
\newblock {\em arXiv preprint arXiv:2507.06261}.

\bibitem[DeepSeek-AI et~al., 2025]{deepseekai2025deepseekv32pushingfrontieropen}
DeepSeek-AI, Liu, A., Mei, A., Lin, B., Xue, B., Wang, B., Xu, B., Wu, B., Zhang, B., Lin, C., Dong, C., Lu, C., Zhao, C., Deng, C., Xu, C., Ruan, C., Dai, D., Guo, D., Yang, D., Chen, D., Li, E., Zhou, F., Lin, F., Dai, F., Hao, G., Chen, G., Li, G., Zhang, H., Xu, H., Li, H., Liang, H., Wei, H., Zhang, H., Luo, H., Ji, H., Ding, H., Tang, H., Cao, H., Gao, H., Qu, H., Zeng, H., Huang, J., Li, J., Xu, J., Hu, J., Chen, J., Xiang, J., Yuan, J., Cheng, J., Zhu, J., Ran, J., Jiang, J., Qiu, J., Li, J., Song, J., Dong, K., Gao, K., Guan, K., Huang, K., Zhou, K., Huang, K., Yu, K., Wang, L., Zhang, L., Wang, L., Zhao, L., Yin, L., Guo, L., Luo, L., Ma, L., Wang, L., Zhang, L., Di, M.~S., Xu, M.~Y., Zhang, M., Zhang, M., Tang, M., Zhou, M., Huang, P., Cong, P., Wang, P., Wang, Q., Zhu, Q., Li, Q., Chen, Q., Du, Q., Xu, R., Ge, R., Zhang, R., Pan, R., Wang, R., Yin, R., Xu, R., Shen, R., Zhang, R., Liu, S.~H., Lu, S., Zhou, S., Chen, S., Cai, S., Chen, S., Hu, S., Liu, S., Hu, S., Ma, S., Wang, S., Yu, S., Zhou,
  S., Pan, S., Zhou, S., Ni, T., Yun, T., Pei, T., Ye, T., Yue, T., Zeng, W., Liu, W., Liang, W., Pang, W., Luo, W., Gao, W., Zhang, W., Gao, X., Wang, X., Bi, X., Liu, X., Wang, X., Chen, X., Zhang, X., Nie, X., Cheng, X., Liu, X., Xie, X., Liu, X., Yu, X., Li, X., Yang, X., Li, X., Chen, X., Su, X., Pan, X., Lin, X., Fu, X., Wang, Y.~Q., Zhang, Y., Xu, Y., Ma, Y., Li, Y., Li, Y., Zhao, Y., Sun, Y., Wang, Y., Qian, Y., Yu, Y., Zhang, Y., Ding, Y., Shi, Y., Xiong, Y., He, Y., Zhou, Y., Zhong, Y., Piao, Y., Wang, Y., Chen, Y., Tan, Y., Wei, Y., Ma, Y., Liu, Y., Yang, Y., Guo, Y., Wu, Y., Wu, Y., Cheng, Y., Ou, Y., Xu, Y., Wang, Y., Gong, Y., Wu, Y., Zou, Y., Li, Y., Xiong, Y., Luo, Y., You, Y., Liu, Y., Zhou, Y., Wu, Z.~F., Ren, Z.~Z., Zhao, Z., Ren, Z., Sha, Z., Fu, Z., Xu, Z., Xie, Z., Zhang, Z., Hao, Z., Gou, Z., Ma, Z., Yan, Z., Shao, Z., Huang, Z., Wu, Z., Li, Z., Zhang, Z., Xu, Z., Wang, Z., Gu, Z., Zhu, Z., Li, Z., Zhang, Z., Xie, Z., Gao, Z., Pan, Z., Yao, Z., Feng, B., Li, H., Cai, J.~L., Ni, J., Xu,
  L., Li, M., Tian, N., Chen, R.~J., Jin, R.~L., Li, S.~S., Zhou, S., Sun, T., Li, X.~Q., Jin, X., Shen, X., Chen, X., Song, X., Zhou, X., Zhu, Y.~X., Huang, Y., Li, Y., Zheng, Y., Zhu, Y., Ma, Y., Huang, Z., Xu, Z., Zhang, Z., Ji, D., Liang, J., Guo, J., Chen, J., Xia, L., Wang, M., Li, M., Zhang, P., Chen, R., Sun, S., Wu, S., Ye, S., Wang, T., Xiao, W.~L., An, W., Wang, X., Sun, X., Wang, X., Tang, Y., Zha, Y., Zhang, Z., Ju, Z., Zhang, Z., and Qu, Z. (2025).
\newblock Deepseek-v3.2: Pushing the frontier of open large language models.

\bibitem[Erdogan et~al., 2025a]{erdogan2025planandactimprovingplanningagents}
Erdogan, L.~E., Lee, N., Kim, S., Moon, S., Furuta, H., Anumanchipalli, G., Keutzer, K., and Gholami, A. (2025a).
\newblock Plan-and-act: Improving planning of agents for long-horizon tasks.

\bibitem[Erdogan et~al., 2025b]{erdogan2025plan-and-act}
Erdogan, L.~E., Lee, N., Kim, S., Moon, S., Furuta, H., Anumanchipalli, G., Keutzer, K., and Gholami, A. (2025b).
\newblock Plan-and-act: Improving planning of agents for long-horizon tasks.
\newblock {\em arXiv preprint arXiv:2503.09572}.

\bibitem[Feng et~al., 2024]{feng2024agilenovelreinforcementlearning}
Feng, P., He, Y., Huang, G., Lin, Y., Zhang, H., Zhang, Y., and Li, H. (2024).
\newblock Agile: A novel reinforcement learning framework of llm agents.

\bibitem[Google, 2025]{kaggleDeepSearchQA}
Google (2025).
\newblock {D}eep{S}earch{Q}{A} --- kaggle.com.
\newblock \url{https://www.kaggle.com/datasets/deepmind/deepsearchqa}.
\newblock [Accessed 05-01-2026].

\bibitem[Guo et~al., 2025]{guo2025deepseek-r1}
Guo, D., Yang, D., Zhang, H., Song, J., Zhang, R., Xu, R., Zhu, Q., Ma, S., Wang, P., Bi, X., et~al. (2025).
\newblock Deepseek-r1: Incentivizing reasoning capability in llms via reinforcement learning.
\newblock {\em arXiv preprint arXiv:2501.12948}.

\bibitem[Han et~al., 2025]{han2025joyagents}
Han, A., Hu, J., Wei, P., Zhang, Z., Guo, Y., Lu, J., and Zhang, Z. (2025).
\newblock Joyagents-r1: Joint evolution dynamics for versatile multi-llm agents with reinforcement learning.
\newblock {\em arXiv preprint arXiv:2506.19846}.

\bibitem[Hu et~al., 2025a]{hu2025stepdeepresearchtechnicalreport}
Hu, C., Du, H., Wang, H., Lin, L., Chen, M., Liu, P., Miao, R., Yue, T., You, W., Ji, W., Yuan, W., Deng, W., Yuan, X., Zhang, X., Liu, X., Liu, X., Xu, Y., Cao, Y., Zhang, Y., Wang, Y., Shu, Y., Zhang, Y., Zhang, Y., Gong, Z., Chang, Z., Li, B., Ma, D., Jia, F., Wang, H., Liu, J., Bai, J., Liu, J., Liu, M., Wang, N., Wu, Q., Du, Q., Li, S., Sun, W., Gong, Y., Chen, Y., Zhao, Y., Lin, Y., Ren, Z., Wang, Z., Zhang, A., Li, B., Ma, B., An, K., Xie, L., Li, M., Li, P., Yang, S., Chen, X., Liu, X., Luo, Y., Song, Y., Ding, Y., Liang, Y., Li, Z., Zhang, Z., Zhang, Z., Jiao, B., Jiang, D., Chen, J., Li, J., Zhang, X., and Zhu, Y. (2025a).
\newblock Step-deepresearch technical report.

\bibitem[Hu et~al., 2025b]{hu2025owloptimizedworkforcelearning}
Hu, M., Zhou, Y., Fan, W., Nie, Y., Xia, B., Sun, T., Ye, Z., Jin, Z., Li, Y., Chen, Q., Zhang, Z., Wang, Y., Ye, Q., Ghanem, B., Luo, P., and Li, G. (2025b).
\newblock Owl: Optimized workforce learning for general multi-agent assistance in real-world task automation.

\bibitem[Hu et~al., 2026a]{hu2026memoryageaiagents}
Hu, Y., Liu, S., Yue, Y., Zhang, G., Liu, B., Zhu, F., Lin, J., Guo, H., Dou, S., Xi, Z., Jin, S., Tan, J., Yin, Y., Liu, J., Zhang, Z., Sun, Z., Zhu, Y., Sun, H., Peng, B., Cheng, Z., Fan, X., Guo, J., Yu, X., Zhou, Z., Hu, Z., Huo, J., Wang, J., Niu, Y., Wang, Y., Yin, Z., Hu, X., Liao, Y., Li, Q., Wang, K., Zhou, W., Liu, Y., Cheng, D., Zhang, Q., Gui, T., Pan, S., Zhang, Y., Torr, P., Dou, Z., Wen, J.-R., Huang, X., Jiang, Y.-G., and Yan, S. (2026a).
\newblock Memory in the age of ai agents.

\bibitem[Hu et~al., 2026b]{hu2026flowsearchadvancingdeepresearch}
Hu, Y., Ma, R., Fan, Y., Shi, J., Cao, Z., Zhou, Y., Yuan, J., Zhang, S., Feng, S., Yan, X., Zhang, S., Zhang, W., Bai, L., and Zhang, B. (2026b).
\newblock Flowsearch: Advancing deep research with dynamic structured knowledge flow.

\bibitem[iQuest, 2025]{iquestlabIQuestCoder}
iQuest (2025).
\newblock {I}{Q}uest {C}oder --- iquestlab.github.io.
\newblock \url{https://iquestlab.github.io/}.
\newblock [Accessed 15-01-2026].

\bibitem[Jin et~al., 2025]{jin2025search}
Jin, B., Zeng, H., Yue, Z., Yoon, J., Arik, S., Wang, D., Zamani, H., and Han, J. (2025).
\newblock Search-r1: Training llms to reason and leverage search engines with reinforcement learning.
\newblock {\em arXiv preprint arXiv:2503.09516}.

\bibitem[Kim et~al., 2024]{kim-etal-2024-rada}
Kim, M., Bursztyn, V., Koh, E., Guo, S., and Hwang, S.-w. (2024).
\newblock {R}a{DA}: Retrieval-augmented web agent planning with {LLM}s.
\newblock In Ku, L.-W., Martins, A., and Srikumar, V., editors, {\em Findings of the Association for Computational Linguistics: ACL 2024}, pages 13511--13525, Bangkok, Thailand. Association for Computational Linguistics.

\bibitem[LangChain, 2025]{githubGitHubLangchainaideepagents}
LangChain (2025).
\newblock {G}it{H}ub - langchain-ai/deepagents: {D}eep {A}gents is an agent harness built on langchain and langgraph. {D}eep {A}gents are equipped with a planning tool, a filesystem backend, and the ability to spawn subagents - making them well-equipped to handle complex agentic tasks. --- github.com.
\newblock \url{https://github.com/langchain-ai/deepagents}.
\newblock [Accessed 15-01-2026].

\bibitem[Li et~al., 2025a]{li2025agentorientedplanningmultiagentsystems}
Li, A., Xie, Y., Li, S., Tsung, F., Ding, B., and Li, Y. (2025a).
\newblock Agent-oriented planning in multi-agent systems.

\bibitem[Li et~al., 2025b]{li2025torl}
Li, X., Zou, H., and Liu, P. (2025b).
\newblock Torl: Scaling tool-integrated rl.
\newblock {\em arXiv preprint arXiv:2503.23383}.

\bibitem[Li et~al., 2025c]{li2025encouraginggoodprocessesneed}
Li, Z., Hu, Y., and Wang, W. (2025c).
\newblock Encouraging good processes without the need for good answers: Reinforcement learning for llm agent planning.

\bibitem[Mialon et~al., 2023]{mialon2023gaia}
Mialon, G., Fourrier, C., Wolf, T., LeCun, Y., and Scialom, T. (2023).
\newblock Gaia: a benchmark for general ai assistants.
\newblock In {\em The Twelfth International Conference on Learning Representations}.

\bibitem[OpenAI, 2025]{openaiIntroducingGPT52}
OpenAI (2025).
\newblock {I}ntroducing {G}{P}{T}-5.2 --- openai.com.
\newblock \url{https://openai.com/index/introducing-gpt-5-2/}.
\newblock [Accessed 08-01-2026].

\bibitem[Paglieri et~al., 2025]{paglieri2025learning}
Paglieri, D., Cupia{\l}, B., Cook, J., Piterbarg, U., Tuyls, J., Grefenstette, E., Foerster, J.~N., Parker-Holder, J., and Rockt{\"a}schel, T. (2025).
\newblock Learning when to plan: Efficiently allocating test-time compute for llm agents.
\newblock {\em arXiv preprint arXiv:2509.03581}.

\bibitem[Parmar et~al., 2025]{parmar2025plangenmultiagentframeworkgenerating}
Parmar, M., Liu, X., Goyal, P., Chen, Y., Le, L., Mishra, S., Mobahi, H., Gu, J., Wang, Z., Nakhost, H., Baral, C., Lee, C.-Y., Pfister, T., and Palangi, H. (2025).
\newblock Plangen: A multi-agent framework for generating planning and reasoning trajectories for complex problem solving.

\bibitem[Qin et~al., 2025]{qin2025flashsearcherfasteffectiveweb}
Qin, T., Chen, Q., Wang, S., Xing, H., Zhu, K., Zhu, H., Shi, D., Liu, X., Zhang, G., Liu, J., Jiang, Y.~E., Gao, X., and Zhou, W. (2025).
\newblock Flash-searcher: Fast and effective web agents via dag-based parallel execution.

\bibitem[Rafailov et~al., 2023]{rafailov2023dpo}
Rafailov, R., Sharma, A., Mitchell, E., Manning, C.~D., Ermon, S., and Finn, C. (2023).
\newblock Direct preference optimization: Your language model is secretly a reward model.
\newblock {\em Advances in Neural Information Processing Systems}, 36:53728--53741.

\bibitem[Schulman et~al., 2017]{schulman2017ppo}
Schulman, J., Wolski, F., Dhariwal, P., Radford, A., and Klimov, O. (2017).
\newblock Proximal policy optimization algorithms.
\newblock {\em arXiv preprint arXiv:1707.06347}.

\bibitem[Shi et~al., 2025a]{shi2025taskcraftautomatedgenerationagentic}
Shi, D., Cao, J., Chen, Q., Sun, W., Li, W., Lu, H., Dong, F., Qin, T., Zhu, K., Liu, M., Yang, J., Zhang, G., Liu, J., Zhang, C., Wang, J., Jiang, Y.~E., and Zhou, W. (2025a).
\newblock Taskcraft: Automated generation of agentic tasks.

\bibitem[Shi et~al., 2025b]{shi2025deepresearchsystematicsurvey}
Shi, Z., Chen, Y., Li, H., Sun, W., Ni, S., Lyu, Y., Fan, R.-Z., Jin, B., Weng, Y., Zhu, M., Xie, Q., Guo, X., Yang, Q., Wu, J., Zhao, J., Tang, X., Ma, X., Wang, C., Mao, J., Ai, Q., Huang, J.-T., Wang, W., Zhang, Y., Yang, Y., Tu, Z., and Ren, Z. (2025b).
\newblock Deep research: A systematic survey.

\bibitem[Shinn et~al., 2023]{reflexion}
Shinn, N., Labash, B., and Gopinath, A. (2023).
\newblock Reflexion: an autonomous agent with dynamic memory and self-reflection.
\newblock {\em arXiv preprint}, abs/2303.11366.

\bibitem[Team et~al., 2025a]{5team2025glm45agenticreasoningcoding}
Team, ., Zeng, A., Lv, X., Zheng, Q., Hou, Z., Chen, B., Xie, C., Wang, C., Yin, D., Zeng, H., Zhang, J., Wang, K., Zhong, L., Liu, M., Lu, R., Cao, S., Zhang, X., Huang, X., Wei, Y., Cheng, Y., An, Y., Niu, Y., Wen, Y., Bai, Y., Du, Z., Wang, Z., Zhu, Z., Zhang, B., Wen, B., Wu, B., Xu, B., Huang, C., Zhao, C., Cai, C., Yu, C., Li, C., Ge, C., Huang, C., Zhang, C., Xu, C., Zhu, C., Li, C., Yin, C., Lin, D., Yang, D., Jiang, D., Ai, D., Zhu, E., Wang, F., Pan, G., Wang, G., Sun, H., Li, H., Li, H., Hu, H., Zhang, H., Peng, H., Tai, H., Zhang, H., Wang, H., Yang, H., Liu, H., Zhao, H., Liu, H., Yan, H., Liu, H., Chen, H., Li, J., Zhao, J., Ren, J., Jiao, J., Zhao, J., Yan, J., Wang, J., Gui, J., Zhao, J., Liu, J., Li, J., Li, J., Lu, J., Wang, J., Yuan, J., Li, J., Du, J., Du, J., Liu, J., Zhi, J., Gao, J., Wang, K., Yang, L., Xu, L., Fan, L., Wu, L., Ding, L., Wang, L., Zhang, M., Li, M., Xu, M., Zhao, M., Zhai, M., Du, P., Dong, Q., Lei, S., Tu, S., Yang, S., Lu, S., Li, S., Li, S., Shuang-Li, Yang, S., Yi,
  S., Yu, T., Tian, W., Wang, W., Yu, W., Tam, W.~L., Liang, W., Liu, W., Wang, X., Jia, X., Gu, X., Ling, X., Wang, X., Fan, X., Pan, X., Zhang, X., Zhang, X., Fu, X., Zhang, X., Xu, Y., Wu, Y., Lu, Y., Wang, Y., Zhou, Y., Pan, Y., Zhang, Y., Wang, Y., Li, Y., Su, Y., Geng, Y., Zhu, Y., Yang, Y., Li, Y., Wu, Y., Li, Y., Liu, Y., Wang, Y., Li, Y., Zhang, Y., Liu, Z., Yang, Z., Zhou, Z., Qiao, Z., Feng, Z., Liu, Z., Zhang, Z., Wang, Z., Yao, Z., Wang, Z., Liu, Z., Chai, Z., Li, Z., Zhao, Z., Chen, W., Zhai, J., Xu, B., Huang, M., Wang, H., Li, J., Dong, Y., and Tang, J. (2025a).
\newblock Glm-4.5: Agentic, reasoning, and coding (arc) foundation models.

\bibitem[Team et~al., 2025b]{kimiteam2025kimik2openagentic}
Team, K., Bai, Y., Bao, Y., Chen, G., Chen, J., Chen, N., Chen, R., Chen, Y., Chen, Y., Chen, Y., Chen, Z., Cui, J., Ding, H., Dong, M., Du, A., Du, C., Du, D., Du, Y., Fan, Y., Feng, Y., Fu, K., Gao, B., Gao, H., Gao, P., Gao, T., Gu, X., Guan, L., Guo, H., Guo, J., Hu, H., Hao, X., He, T., He, W., He, W., Hong, C., Hu, Y., Hu, Z., Huang, W., Huang, Z., Huang, Z., Jiang, T., Jiang, Z., Jin, X., Kang, Y., Lai, G., Li, C., Li, F., Li, H., Li, M., Li, W., Li, Y., Li, Y., Li, Z., Li, Z., Lin, H., Lin, X., Lin, Z., Liu, C., Liu, C., Liu, H., Liu, J., Liu, J., Liu, L., Liu, S., Liu, T.~Y., Liu, T., Liu, W., Liu, Y., Liu, Y., Liu, Y., Liu, Y., Liu, Z., Lu, E., Lu, L., Ma, S., Ma, X., Ma, Y., Mao, S., Mei, J., Men, X., Miao, Y., Pan, S., Peng, Y., Qin, R., Qu, B., Shang, Z., Shi, L., Shi, S., Song, F., Su, J., Su, Z., Sun, X., Sung, F., Tang, H., Tao, J., Teng, Q., Wang, C., Wang, D., Wang, F., Wang, H., Wang, J., Wang, J., Wang, J., Wang, S., Wang, S., Wang, Y., Wang, Y., Wang, Y., Wang, Y., Wang, Y., Wang, Z.,
  Wang, Z., Wang, Z., Wei, C., Wei, Q., Wu, W., Wu, X., Wu, Y., Xiao, C., Xie, X., Xiong, W., Xu, B., Xu, J., Xu, J., Xu, L.~H., Xu, L., Xu, S., Xu, W., Xu, X., Xu, Y., Xu, Z., Yan, J., Yan, Y., Yang, X., Yang, Y., Yang, Z., Yang, Z., Yang, Z., Yao, H., Yao, X., Ye, W., Ye, Z., Yin, B., Yu, L., Yuan, E., Yuan, H., Yuan, M., Zhan, H., Zhang, D., Zhang, H., Zhang, W., Zhang, X., Zhang, Y., Zhang, Y., Zhang, Y., Zhang, Y., Zhang, Y., Zhang, Y., Zhang, Z., Zhao, H., Zhao, Y., Zheng, H., Zheng, S., Zhou, J., Zhou, X., Zhou, Z., Zhu, Z., Zhuang, W., and Zu, X. (2025b).
\newblock Kimi k2: Open agentic intelligence.

\bibitem[Team et~al., 2025c]{tongyideepresearchteam2025tongyideepresearchtechnicalreport}
Team, T.~D., Li, B., Zhang, B., Zhang, D., Huang, F., Li, G., Chen, G., Yin, H., Wu, J., Zhou, J., Li, K., Su, L., Ou, L., Zhang, L., Xie, P., Ye, R., Yin, W., Yu, X., Wang, X., Wu, X., Chen, X., Zhao, Y., Zhang, Z., Tao, Z., Zhang, Z., Qiao, Z., Wang, C., Yu, D., Fu, G., Shen, H., Yang, J., Lin, J., Zhang, J., Zeng, K., Yang, L., Yin, H., Song, M., Yan, M., Liao, M., Xia, P., Xiao, Q., Min, R., Ding, R., Fang, R., Chen, S., Huang, S., Wang, S., Cai, S., Shen, W., Wang, X., Guan, X., Geng, X., Shi, Y., Wu, Y., Chen, Z., Li, Z., and Jiang, Y. (2025c).
\newblock Tongyi deepresearch technical report.

\bibitem[Wang et~al., 2024a]{wang2024qimprovingmultistepreasoning}
Wang, C., Deng, Y., Lyu, Z., Zeng, L., He, J., Yan, S., and An, B. (2024a).
\newblock Q*: Improving multi-step reasoning for llms with deliberative planning.

\bibitem[Wang et~al., 2025a]{wang2025openhandsopenplatformai}
Wang, X., Li, B., Song, Y., Xu, F.~F., Tang, X., Zhuge, M., Pan, J., Song, Y., Li, B., Singh, J., Tran, H.~H., Li, F., Ma, R., Zheng, M., Qian, B., Shao, Y., Muennighoff, N., Zhang, Y., Hui, B., Lin, J., Brennan, R., Peng, H., Ji, H., and Neubig, G. (2025a).
\newblock Openhands: An open platform for ai software developers as generalist agents.

\bibitem[Wang et~al., 2024b]{wang2024describeexplainplanselect}
Wang, Z., Cai, S., Chen, G., Liu, A., Ma, X., and Liang, Y. (2024b).
\newblock Describe, explain, plan and select: Interactive planning with large language models enables open-world multi-task agents.

\bibitem[Wang et~al., 2025b]{wang2025ragen}
Wang, Z., Wang, K., Wang, Q., Zhang, P., Li, L., Yang, Z., Jin, X., Yu, K., Nguyen, M.~N., Liu, L., et~al. (2025b).
\newblock Ragen: Understanding self-evolution in llm agents via multi-turn reinforcement learning.
\newblock {\em arXiv preprint arXiv:2504.20073}.

\bibitem[Wei et~al., 2022]{cot}
Wei, J., Wang, X., Schuurmans, D., Bosma, M., Ichter, B., Xia, F., Chi, E., Le, Q., and Zhou, D. (2022).
\newblock Chain-of-thought prompting elicits reasoning in large language models.

\bibitem[Wolfson et~al., 2026]{wolfson2026monaco}
Wolfson, T., Trivedi, H., Geva, M., Goldberg, Y., Roth, D., Khot, T., Sabharwal, A., and Tsarfaty, R. (2026).
\newblock Monaco: More natural and complex questions for reasoning across dozens of documents.
\newblock {\em Transactions of the Association for Computational Linguistics}, 14:23--46.

\bibitem[Wu et~al., 2025a]{wu2025webwalkerbenchmarkingllmsweb}
Wu, J., Yin, W., Jiang, Y., Wang, Z., Xi, Z., Fang, R., Zhang, L., He, Y., Zhou, D., Xie, P., and Huang, F. (2025a).
\newblock Webwalker: Benchmarking llms in web traversal.

\bibitem[Wu et~al., 2025b]{wu2025gapgraphbasedagentplanning}
Wu, J., Zhao, Q., Chen, Z., Qin, K., Zhao, Y., Wang, X., and Yao, Y. (2025b).
\newblock Gap: Graph-based agent planning with parallel tool use and reinforcement learning.

\bibitem[Xi et~al., 2025]{xi2025agentgymrltrainingllmagents}
Xi, Z., Huang, J., Liao, C., Huang, B., Guo, H., Liu, J., Zheng, R., Ye, J., Zhang, J., Chen, W., He, W., Ding, Y., Li, G., Chen, Z., Du, Z., Yao, X., Xu, Y., Chen, J., Gui, T., Wu, Z., Zhang, Q., Huang, X., and Jiang, Y.-G. (2025).
\newblock Agentgym-rl: Training llm agents for long-horizon decision making through multi-turn reinforcement learning.

\bibitem[Yang et~al., 2025]{yang2025qwen3technicalreport}
Yang, A., Li, A., Yang, B., Zhang, B., Hui, B., Zheng, B., Yu, B., Gao, C., Huang, C., Lv, C., Zheng, C., Liu, D., Zhou, F., Huang, F., Hu, F., Ge, H., Wei, H., Lin, H., Tang, J., Yang, J., Tu, J., Zhang, J., Yang, J., Yang, J., Zhou, J., Zhou, J., Lin, J., Dang, K., Bao, K., Yang, K., Yu, L., Deng, L., Li, M., Xue, M., Li, M., Zhang, P., Wang, P., Zhu, Q., Men, R., Gao, R., Liu, S., Luo, S., Li, T., Tang, T., Yin, W., Ren, X., Wang, X., Zhang, X., Ren, X., Fan, Y., Su, Y., Zhang, Y., Zhang, Y., Wan, Y., Liu, Y., Wang, Z., Cui, Z., Zhang, Z., Zhou, Z., and Qiu, Z. (2025).
\newblock Qwen3 technical report.

\bibitem[Yang et~al., 2024]{yang2024sweagentagentcomputerinterfacesenable}
Yang, J., Jimenez, C.~E., Wettig, A., Lieret, K., Yao, S., Narasimhan, K., and Press, O. (2024).
\newblock Swe-agent: Agent-computer interfaces enable automated software engineering.

\bibitem[Yao et~al., 2023a]{tot}
Yao, S., Yu, D., Zhao, J., Shafran, I., Griffiths, T.~L., Cao, Y., and Narasimhan, K. (2023a).
\newblock Tree of thoughts: Deliberate problem solving with large language models.

\bibitem[Yao et~al., 2023b]{yao2023react}
Yao, S., Zhao, J., Yu, D., Du, N., Shafran, I., Narasimhan, K.~R., and Cao, Y. (2023b).
\newblock React: Synergizing reasoning and acting in language models.
\newblock In {\em The Eleventh International Conference on Learning Representations}.

\bibitem[Zhang et~al., 2025]{zhang2025cosightenhancingllmbasedagents}
Zhang, H., Lu, J., Jiang, S., Zhu, C., Xie, L., Zhong, C., Chen, H., Zhu, Y., Du, Y., Gao, Y., Huang, L., Wang, B., Tan, F., and Zou, P. (2025).
\newblock Co-sight: Enhancing llm-based agents via conflict-aware meta-verification and trustworthy reasoning with structured facts.

\bibitem[Zhou et~al., 2023]{zhou2023language}
Zhou, A., Yan, K., Shlapentokh-Rothman, M., Wang, H., and Wang, Y.-X. (2023).
\newblock Language agent tree search unifies reasoning acting and planning in language models.
\newblock {\em arXiv preprint arXiv:2310.04406}.

\bibitem[Zhu et~al., 2025]{zhu2025oagentsempiricalstudybuilding}
Zhu, H., Qin, T., Zhu, K., Huang, H., Guan, Y., Xia, J., Yao, Y., Li, H., Wang, N., Liu, P., Peng, T., Gui, X., Li, X., Liu, Y., Jiang, Y.~E., Wang, J., Zhang, C., Tang, X., Zhang, G., Yang, J., Liu, M., Gao, X., Liu, J., and Zhou, W. (2025).
\newblock Oagents: An empirical study of building effective agents.

\end{thebibliography}
